\documentclass{sig-alternate}
\pdfoutput=1
\usepackage{subfigure, multirow, times}
\usepackage[ruled,vlined,linesnumbered]{algorithm2e}

\usepackage{amsfonts, amsmath,bm}
\usepackage{mdwmath}
\usepackage{mathrsfs}
\usepackage{graphicx}

\usepackage{amsmath, amsfonts, amssymb, xspace, color}

\newcommand{\hide}[1]{} 
\newcommand{\vpara}[1]{\vspace{0.08in}\noindent\textbf{#1}}

\newcommand{\pa}[2]{\vspace{#1}\noindent\textbf{#2}} 
\newcommand{\ra}[1]{\overset{#1}{\longrightarrow}}

\newtheorem{definition}{Definition}

\newtheorem{example}{Example}

\def \c {\mathbf{c}}

\def \r {\mathbf{r}}

\def \v {\mathbf{v}}

\def \x {\mathbf{x}}

\def \L {\mathbf{L}}
\def \P {\mathbf{P}}

\def \C {\mathcal{C}}
\def \D {\mathcal{D}}

\def \S {\mathcal{S}}
\def \T {\mathcal{T}}

\begin{document}

\title{p-Causality: Identifying Spatiotemporal Causal Pathways for Air Pollutants with Urban Big Data}

 \author{
 \alignauthor
 Julie Yixuan Zhu$^{1,3,*}$ ~~  Chao Zhang$^{2,3,*\thanks{\small{*Equal contribution. The first authors were interns supervised by the correspondence author in MSRA.}}}$ ~~  Huichu Zhang$^{2,4}$ ~~ Shi Zhi$^2$ ~~ Victor O.K. Li$^1$ ~~ \\Jiawei Han$^2$ ~~ Yu Zheng$^{3,+\thanks{\small{+Correspondence author.}}}$\\
        $^1$\affaddr{Department of Electrical and Electronic Engineering, the University of Hong Kong, HK}\\
        $^2$\affaddr{Dept. of Computer Science, University of Illinois at Urbana-Champaign, Urbana, IL, USA}\\
        $^3$\affaddr{Microsoft Research Asia, Beijing, China}\\
		$^4$\affaddr{Apex Data \& Knowledge Management Lab, Shanghai Jiao Tong University, Shanghai.}\\
        \email{$^1$\{yxzhu,vli\}@eee.hku.hk ~~~$^2$\{czhang82,shizhi2, hanj\}@illinois.edu ~~~~~$^3$yuzheng@microsoft.com ~~~$^4$zhc@apex.sjtu.edu.cn}
 }
\renewcommand\footnotemark{}
\renewcommand\footnoterule{}

\maketitle

\begin{abstract}

Many countries are suffering from severe air pollution.
Understanding how different air pollutants accumulate and propagate is critical to making relevant public policies. 
In this paper, we use urban big data (air quality data and meteorological data) to identify the
\emph{spatiotemporal (ST) causal pathways} for air pollutants. 
This problem is challenging because: (1) there are numerous noisy and low-pollution periods in
the raw air quality data, which may lead to unreliable causality analysis; 
(2) for large-scale data in the ST space, the computational complexity of constructing a causal structure is very high; 
and (3) the \emph{ST causal pathways} are complex due to 
the interactions of multiple pollutants and the influence of environmental
factors. Therefore, we present \emph{pg-Causality}, a novel pattern-aided graphical causality analysis approach that combines the strengths of \emph{pattern
mining} and \emph{Bayesian learning} to efficiently identify the \emph{ST causal pathways}. First, \emph{pattern
mining} helps suppress the noise by capturing frequent evolving patterns (FEPs) of each monitoring sensor, 
and greatly reduce the complexity by selecting the pattern-matched sensors as ``causers''. 
Then, \emph{Bayesian learning} carefully encodes
the local and ST causal relations with a Gaussian Bayesian Network (GBN)-based graphical model, 
which also integrates environmental influences to minimize biases in the final results. 
We evaluate our approach with three real-world data sets containing 982 air quality sensors in 128 cities, in
three regions of China from 01-Jun-2013 to 31-Dec-2016. Results show that our
approach outperforms the traditional causal structure learning methods in time efficiency, inference accuracy and interpretability.

\end{abstract}



\section{Introduction}
\label{sect:intro}

Recent years have witnessed the air pollution problem becoming a severe
environmental and societal issue around the world.  For example, in 2015, the average concentration of PM2.5 in Beijing
is greater than 150, classified as hazardous to human health by the World
Health Organization, on more than 46 days. On Dec 7th 2015, the Chinese
government issues the first red alert because of the extremely heavy air
pollution, leading to suspended schools, closed construction sites, and traffic
restrictions. Though many ways have been deployed to reduce the air pollution,
the severe air pollution in Beijing has not been significantly alleviated.

Identifying the causalities has become an urgent
problem for mitigating the air pollution and suggesting relevant public policy making. 
Previous research on the air pollution cause identification mostly relies on chemical
receptor \cite{lee2008source} or dispersion models
\cite{keats2007dispersion}. However, these approaches often involve 
domain-specific data collection which is labor-intensive, or require theoretical assumptions that real-world data may not guarantee.
Recently, with the increasingly available air quality data
collected by versatile sensors deployed in different regions, and pubic
meteorological data, it is possible to analyze the causality of air pollution
through a data-driven approach.

The goal of our research is to learn the \emph{spatiotemporal (ST) causal pathways} among different
pollutants, by mining the dependencies among
air pollutants under different environmental influences.  Fig.
\ref{fig:example} shows two example causal pathways for PM10 in Beijing. Let
us first consider the pathway in Fig. \ref{fig:example}(a). When the wind
speed is less than 5 m/s, the high concentration of PM10 in Beijing is mainly caused by SO$_2$ in
Zhangjiakou and PM2.5 in Baoding. In contrast, as shown in Fig.
\ref{fig:example}(b), when the wind speed is larger than 5m/s, PM10 in Beijing is mainly due to PM2.5
in Zhangjiakou and NO$_2$ in Chengde. Based on this example, we can see the \emph{spatiotemporal (ST) causal pathways} should reflect the following two aspects: 1) the
\emph{structural dependency}, which indicates the reactions and propagations of
multiple pollutants in the ST space; and 2) the \emph{global confounder}, which
denotes how different environmental conditions could lead to different
causal pathways.

\begin{figure}[h]
\label{fig:example}
\centering
\includegraphics[width=3.3in]{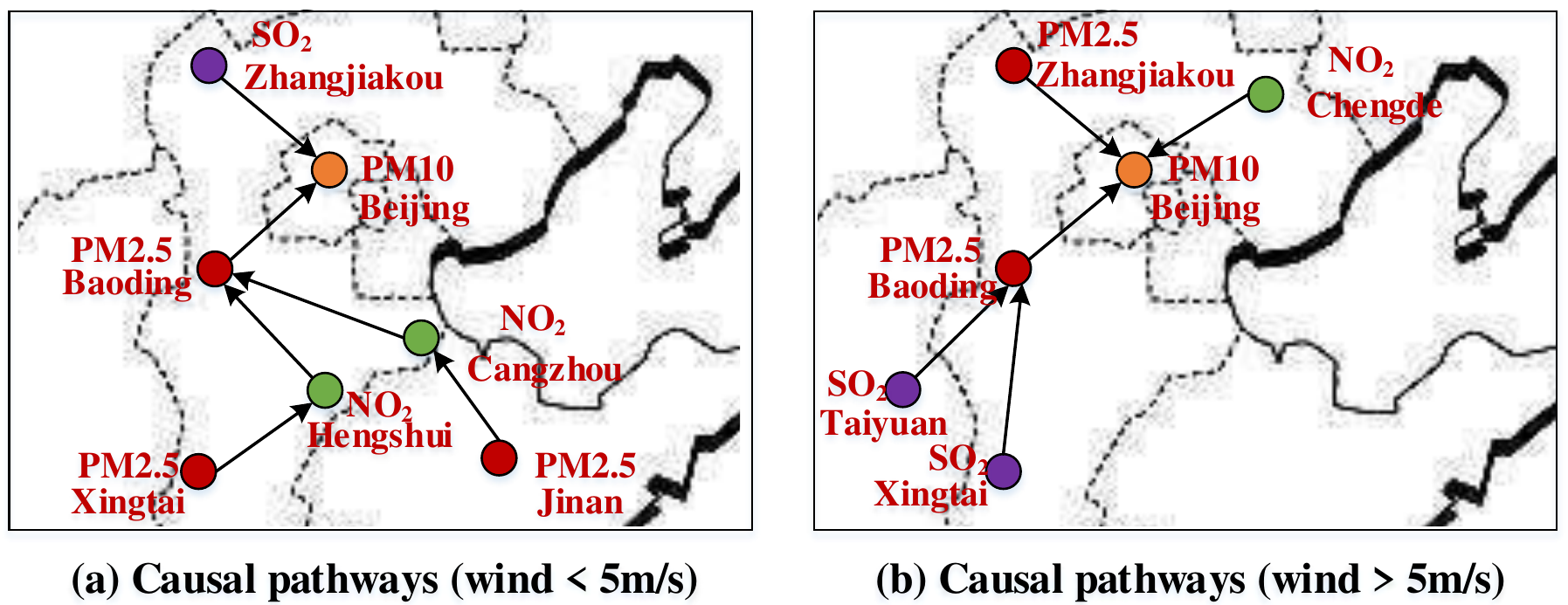}

\caption{\label{fig:example} An illustration of identifying causal pathways.}
\end{figure}


However, identifying the \emph{ST causal pathways} from big air quality and meteorological data is
not trivial because of the following challenges.  \emph{First}, not all air
pollution data are useful for causality analysis. In the raw sensor-collected
air quality data, there are numerous uninteresting fluctuations and noisy
variations. Including such data into the causality analysis process is expected
to lead to unreliable conclusions.  \emph{Second}, the sheer size of the air
quality makes the causality analysis difficult.  In most air quality monitoring
applications, thousands of sensors are deployed at different locations to record
the air quality hourly for years.  Discovering the ST causal relationships from
such a large scale is challenging.  \emph{Third}, air pollution causal
pathways are complex in nature. The air polluting process typically involves
multiple types of pollutants that are mutually interacting, and is subject to
local reactions, ST propagations and confounding factors, such as wind and
humidity.

Existing data mining techniques for learning the causal pathways have been proposed from two perspectives: pattern-based \cite{zhang2015assembler}\cite{liuwei2015traffic} and Bayesian-based 
\cite{pearl2003causality}\cite{zhu2016gcm}. Pattern-based approaches aim to extract frequently occurring
phenomena from historical data by applying pattern mining techniques; while
Bayesian-based techniques use directed acyclic graphs (DAGs) to encode the causality and then learn the probabilistic dependencies from historical data.
Though inspiring results have been obtained by pattern-based and
Bayesian-based techniques, both approaches have their merits and
downsides. Pattern-based approaches can fast extract a set of patterns
(e.g., frequent patterns, contrast patterns) from historical air quality
data. Such patterns can capture the intrinsic regularity present in historical
air quality data. However, they only provide shallow understanding of the air
polluting process, and there are usually a huge number of frequent patterns,
which largely limits the usability of the pattern set.  On the other hand,
Bayesian-based approaches depict the causal dependencies between multiple air pollutants
in a principled way.  However, the performance of Bayesian-based models is highly
dependent on the quality of the training data. When there exist massive noise
and data sparsity, as the case of the air quality data, the performance of
the Bayesian-based models is limited. 
Besides, Bayesian-based approaches are limited by high computational cost \cite{chickering1996} and the impact of confounding \cite{sun2015causal}.

We propose \emph{pg-Causality}, which combines \emph{pattern mining} with \emph{Bayesian learning} to unleash the strengths of both. We claim \emph{pg-Causality} 
is essential for \emph{ST causal pathway} identification, with the contributions listed as below:

$\bullet$ First, we propose a framework that combines frequent pattern mining with Bayesian-based graphical model to identify the spatiotemporal (ST) causal relationship between air pollutants in the ST space. The frequent pattern mining \cite{zhang2014splitter} can accurately estimate the correlation between the air quality of each pair of locations, capturing the meaningful fluctuation of two time series. Using the correlation patterns, whose scales are significantly smaller than the raw data, as an input of a Bayesian network (BN), the computational complexity of the Bayesian network causality model has been significantly reduced. The patterns also help suppress the noise for learning a Bayesian network's structure. This not only leads to a more efficient but also more effective causal pathway identification. We also integrate the environmental factors in the Bayesian-based graphical model to minimize the biases in the final results.


$\bullet$ Second, we have carefully evaluated our proposed approach on three real data sets
with 3.5 years' air quality and meteorological data collected
from hundreds of cities in China. Our results show that the proposed approach
is significantly better than the existing baseline methods in time efficiency,
inference accuracy and interpretability.

\section{Framework}
In this section, we first describe the problem of identifying spatio-temporal
causal pathways for air pollutants, and then introduce the framework of \emph{pg-Causality}.

Let $\S = \{s_1, s_2, \ldots, s_n,\dots\}$ be the location set of the
air quality monitoring sensors deployed in a geographical region. Each sensor
is deployed at a location $s_n \in \S$
to periodically measure the target condition around it. All sensors have
synchronized measurements over the time domain $ \T = \{ 1, 2, \ldots,
\mathrm{T} \} $, where each $t\in\T$ is a timestamp.  We also consider a set
$\C = \{c_1, c_2, \ldots, c_\mathrm{M}\}$ of pollutants. Given $c_m \in \C$,
$s_n \in \S$, and $t \in \T$ ($1 \le m \le \mathrm{M}, 1 \le n \le \mathrm{N},
1 \le t \le \mathrm{T}$), we use $P_{c_m s_n t}$ to denote the measurement of
pollutant $c_m$ at location $s_n$ and timestamp $t$. In addition, we also have
the meteorological data at timestamp $t$ for the entire geographical region,
denoted as $\bm{E_t}$, as a vector of environmental factors. 
Using the air pollutant measurements and meteorological data, we aim to
identify faithful causal relationships among different pollutants at different
locations. We integrated the environmental facotors $\bm{E_t}$ to the causal pathways through a graphical model, 
setting the number of clusters as $\mathrm{K}$ and time lag constraint as $\mathrm{L}$. We list the notations in TABLE 1.

\begin{table}[!htb]
\centering
\vspace{-3ex}
\caption{Notation Table.}
    \label{tab:notation}

    \begin{small}
    \begin{tabular}{|c|l|}
    \hline
    $\S$ & The location set of the air quality monitoring sensors. 
	\\ & $\S = \{s_1, s_2, \ldots, s_n,\dots\}$\\\hline
    $s_n \in \S$ & The location of the $n$-th neighborhood sensor. \\\hline
	$s_0$ & The location of the target sensor. \\\hline
	$\mathrm{N}$ & Number of ``causers'' in the neighborhood. \\\hline
    $\T$ & Timestamps domain $\T = \{ 1, 2, \ldots, \mathrm{T}\}$. \\\hline
    $t\in\T$ & The current timestamp. \\\hline
	$\mathrm{T}$ & Number of timestamps. \\\hline
    $\C$ & Category set of pollutants $\C = \{c_1, c_2, \ldots, c_\mathrm{M}\}$. \\\hline
	$\mathrm{M}$ & Number of pollutants measured by each sensor. \\\hline
	$c_m \in \C$ & The pollutant of the $m$-th category. \\\hline
	$c_{m_n}$ & The most likely category of ``causer'' pollutant at $s_n$. \\\hline
	$P_{c_m s_n t}$ & Pollutant $c_m$ at location $s_n$ and timestamp $t$. 
	\\ & $1 \le m \le \mathrm{M}, 1 \le n \le \mathrm{N}, 1 \le t \le \mathrm{T}$.\\\hline
	$\mathrm{K}$ & Number of clusters in the graphical causality model. \\\hline
	$l\in[1,\mathrm{L}]$ & Time lag in the graphical causality model. \\\hline
	$\bm{E_t}$ & The environmental factors. $\bm{E_t}=\{E_t^{(1)},E_t^{(2)},\dots\}$. \\\hline
    \end{tabular}
    \end{small}
\end{table}

Fig. \ref{fig:framework} shows the framework of our proposed
approach \emph{pg-Causality}. It consists of two main
modules: pattern mining and Bayesian Network Learning, detailed as follows. 

\begin{figure}[!htb]
\label{fig:framework}
\centering
\includegraphics[width=2.8in]{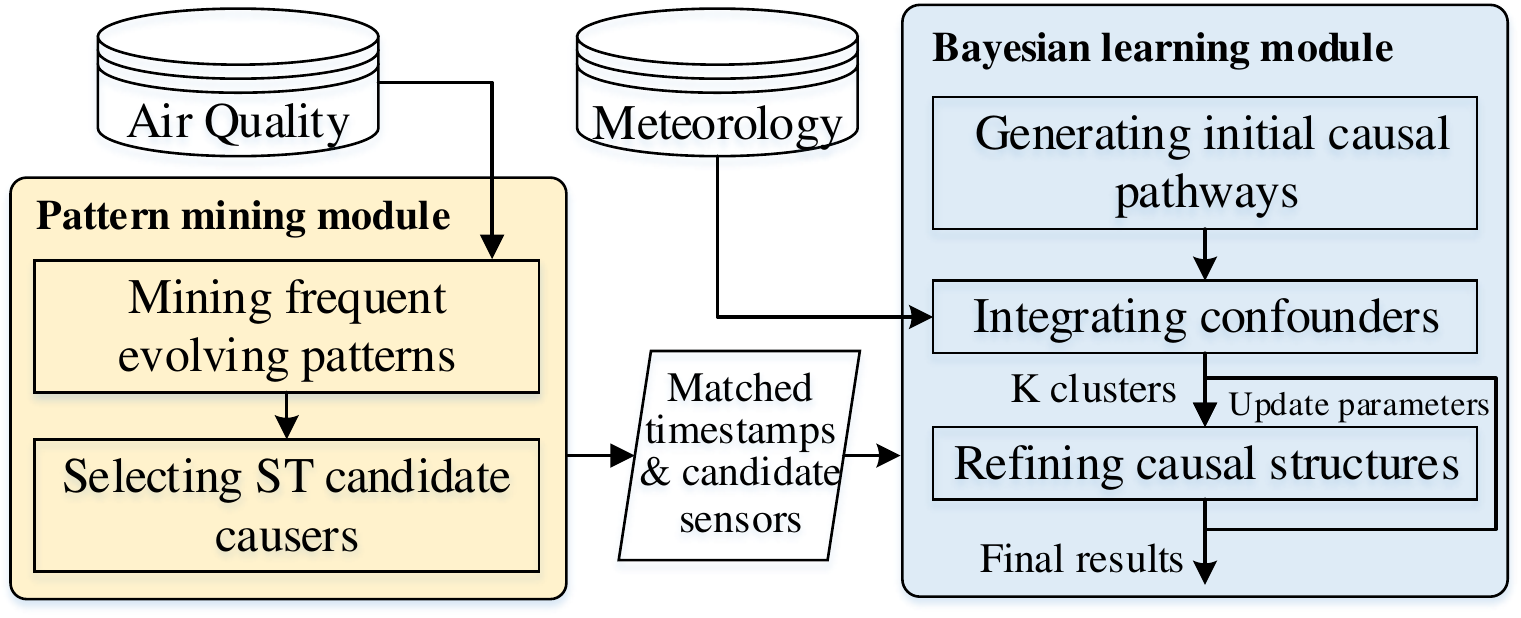}

\caption{\label{fig:framework} The framework of our approach.}
\end{figure}

\vpara{Pattern Mining Module:}  This module first extracts the \emph{frequent
evolving patterns} (FEPs) \cite{zhang2014splitter} for each sensor. The FEPs essentially capture the air
quality changing behaviors that frequently appear on the target sensor. By
mining all FEPs from the historical air quality data, this module
efficiently captures the regularity in raw data and largely reduces the noise (Section \ref{sect:def-pattern} and \ref{sect:pattern-algo}). Afterwards, we examine the pattern-based similarities between locations to select candidate
causers for each target sensor. By comparing the FEPs
occurring on different sensors, we can obtain a shallow understanding of the
causal relationships between different sensors, which can be further utilized
to simplify learning the causal structures (Section \ref{subsect:cand}).

\vpara{Bayesian Learning Module:} By using the matched timestamps of the extracted FEPs at different sensors, together with the selected candidate sensors in the pattern mining module, this module further trains high-quality causal pathways from the large-scale air quality and context data in an effective and scalable way. We first generate the initial causal pathways from the selected candidate causers, taking into account both the local interactions of multiple air pollutants and the ST propagations (Section \ref{sect:step3}). Then to minimize the impact of confounding (Section \ref{sect:step4}), we integrate the confounders (e.g., wind, humidity) into the a Gaussian Bayesian Network (GBN)-based graphical model. Last, we refine the parameters and structures of the Bayesian network to generate the final causal pathways (Section \ref{sect:step5}).


We argue that the combination of two modules helps efficiently identify the causal pathways of the air pollutants. First, the meaningful behaviors of each time series selected by the pattern mining module could significantly reduce the noise in calculating the causal relationships. For example, Fig. \ref{fig:sensors}(a) shows an illustration of three time series at sensors $1$, $2$, and $3$, in North China, with sensor $1$ as the target sensor. When simply using statistical models to identify the dependencies among the three time series, the causal pathway $2\rightarrow1$ and $3\rightarrow1$ cannot be faithfully justified, since the fluctuations and low pollution periods will make the dependency metric for sensors $2\rightarrow1$ and $3\rightarrow1$ very similar. By using the pattern mining module, we found that the increasing behaviors of sensor $2$ frequently happen before sensor $1$, and thus can select sensor $2$ as the candidate ``causer'' for target sensor $1$. 
Second, the selected ``causers'' by the pattern mining module will greatly reduce the complexity of the Bayesian structure learning. For example, Fig. \ref{fig:sensors}(b) illlustrates a scenario of learning the 1-hop Bayesian structure from 100 sensors to a target pollutant. We use the pattern mining module to select top ``N = 2'' candidate causers, thus reducing the searching space from $O(100)$ to $O(2)$ for Bayesian structure construcion. Third, we verify the effectiveness of causal pathway learning with pg-Causality, compared with only using Bayesian learning without pattern mining. Combining pattern mining with Bayesian learning demonstrates better inference accuracy, time efficiency, and interpretability.

\begin{figure}[!htb]
\label{fig:sensors}
\centering

\includegraphics[width=3.4in]{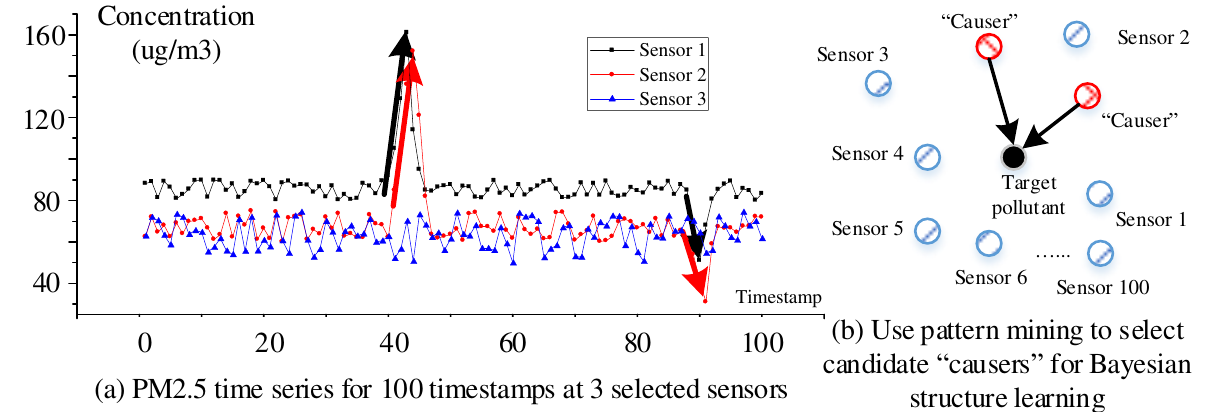}

\caption{\label{fig:sensors} Illustration of how pattern mining helps to reduce the effect of fluctuations in causal structure learning.}
\end{figure}

\section{The Pattern Mining Module}
\label{sect:pattern}


\subsection{Frequent Evolving Pattern}
\label{sect:def-pattern}

To capture frequent evolving behaviors of each sensor, we define
\emph{frequent evolving pattern} (FEP), an adaption of the classic sequential
pattern concept \cite{PeiHPCDH01}. As the sequential patterns are defined on
transactional sequences, we first discretize the raw air quality data. Given a
pollutant $c_m$ at sensor $s_n$, the measurements of $c_m$ at $s_n$ over the
time domain $\T$ form a time series. We discretize the time series as follows:
(1) partition it by day to obtain a collection of daily time series, denoted as
$ P_{c_m s_n} $; and (2) for each daily time series $\langle({p}_1, t_1),
({p}_2,t_2), \ldots, ({p}_l,t_l)\rangle$, map every real-value measure $p_i$
($1 \le i \le l$) to a discrete level $\hat{p}_i$ using \emph{symbolic
  approximation aggregation} \cite{lin2003symbolic}. After discretization, we
obtain a database of symbolic sequences, as defined in Definition \ref{def:db}.

\begin{definition} [Symbolic Pollution Database]
  \label{def:db}
  For pollutant $c_m$ and sensor $s_n$, the symbolic pollution database
  $\hat{P}_{c_m s_n}$ is a collection of daily sequences. Each sequence $d \in
  \hat{P}_{c_m s_n}$ has the form $\langle(\hat{p}_1, t_1), (\hat{p}_2,t_2),
  \ldots, (\hat{p}_l,t_l)\rangle$ where an element $(\hat{p}_i, t_i)$ means
  the pollution level of $c_m$ at sensor $s_n$ and time $t_i$ is $\hat{p}_i$.
\end{definition}

Given the database $\hat{P}_{c_m s_n}$, our goal is to find frequent
evolving behaviors of $s_n$ regarding $c_m$. Below, we introduce the concepts of
\emph{evolving sequence} and \emph{occurrence}.

\begin{definition} [Evolving Sequence]
\label{def:sequence}
A length-$k$ evolving sequence $T$ has the form
$
T = \hat{p}_1 \ra{\Delta t} \hat{p}_2 \ra{\Delta t} \cdots \ra{\Delta t} \hat{p}_k
$,
where (1) $\forall i > 1, \hat{p}_{i-1} \ne \hat{p}_i$ and (2) $\Delta t$ is the maximum
transition time between consecutive records.
\end{definition}

\begin{definition} [Occurrence]
\label{def:contain}
Given a daily sequence $d = \langle(\hat{p}_1, t_1), (\hat{p}_2,t_2), \ldots,
(\hat{p}_l,t_l)\rangle$ and an evolving sequence $T = \hat{p}_1 \ra{\Delta t}
\hat{p}_2 \cdots \ra{\Delta t} \hat{p}_k$  ($k \le l$), $T$ occurs in $d$
(denoted as $T \sqsubseteq d$) if there exist integers $1 \le j_1 < j_2 <
\cdots < j_k \le l$ such that: (1) $\forall 1 \le i \le k$, $\hat{p}_{j_i} =
\hat{p}_i$; and (2) $\forall 1 \le i \le k-1$, $0 < t_{j_{i+1}} - t_{j_i} \le \Delta t$.
\end{definition}

For clarity, we denote an evolving sequence $\hat{p}_1 \ra{\Delta t} \hat{p}_2
\cdots \ra{\Delta t} \hat{p}_k$ as $\hat{p}_1 \rightarrow \hat{p}_2 \cdots
\rightarrow \hat{p}_k$ when the context is clear. Now, we proceed to define
\emph{support} and \emph{frequent evolving pattern}.

\begin{definition} [Support]
\label{def:support}
Given $\hat{P}_{c_m s_n}$ and an evolving sequence $T$, the support of $T$ is
the number of days that $T$ occurs, \emph{i.e.}, $Sup(T) = |\{o|o \in
\hat{P}_{c_m s_n} \wedge T \sqsubseteq o\}|$.
\end{definition}

\begin{definition} [Frequent Evolving Pattern]
\label{def:pattern}
Given a support threshold $\sigma$, an evolving sequence $T$ is a frequent evolving pattern in
database $\hat{P}_{c_m s_n}$ if $Sup(T) \ge \sigma$.
\end{definition}

\vspace{-15 pt}

\subsection{The FEP Mining Algorithm}
\label{sect:pattern-algo}

Now we proceed to discuss how to mine all FEPs in any symbolic pollution
database.  It is closely related to the classic sequential pattern mining
problem.  However, recall that there are two constraints in the definition of
FEP: (1) the consecutive symbols must be different; and (2) the time gap
between consecutive records should be no greater than the temporal constraint
$\Delta t$.  A sequential pattern mining algorithm needs to be tailored to
ensure these two constraints are satisfied.

We adapt PrefixSpan \cite{PeiHPCDH01} as it has proved to be one of the most
efficient sequential pattern mining algorithms.  The basic idea of PrefixSpan
is to use short patterns as the prefix to project the database and progressively
grow the short patterns by searching for local frequent items. For a short
pattern $\beta$, the $\beta$-projected database $\D_\beta$ includes the
postfix from the sequences that contain $\beta$. Local frequent items in
$\D_\beta$ are then identified and appended to $\beta$ to form longer patterns.
Such a process is repeated recursively until no more local frequent items
exist. One can refer to \cite{PeiHPCDH01} for more details.

Given a sequence $\alpha$ and a frequent item $\hat{p}$, when creating $\hat{p}$-projected
database, the standard PrefixSpan procedure generates one postfix based on the
first occurrence of $\hat{p}$ in $\alpha$. This strategy, unfortunately, can miss
FEPs in our problem.

\begin{table}[h]
\centering
\vspace{-3ex}
\caption{An example symbolic pollution database.}
	
    \label{tab:conversion}
    \begin{small}
    \begin{tabular}{|c|l|}
    \hline
    {\bfseries Day}    & {\bfseries Daily sequence} \\ \hline \hline
    $d_1$ & $\langle(\hat{p}_2,0),(\hat{p}_1,10),(\hat{p}_2,30),(\hat{p}_3,40)\rangle$ \\\hline
    $d_2$ & $\langle(\hat{p}_1,0),(\hat{p}_2,30),(\hat{p}_1,360),(\hat{p}_2,400),(\hat{p}_3,420)\rangle$ \\\hline
    $d_3$ & $\langle(\hat{p}_2,0),(\hat{p}_3,30)\rangle$ \\\hline
    $d_4$ & $\langle(\hat{p}_1,0),(\hat{p}_1,120),(\hat{p}_3,140),(\hat{p}_2,150),(\hat{p}_3,180)\rangle$ \\\hline
    $d_5$ & $\langle(\hat{p}_2,50),(\hat{p}_2,80),(\hat{p}_3,120), (\hat{p}_1,210)\rangle$ \\\hline
    \end{tabular}
    \end{small}
\end{table}

\begin{example}
Let $\Delta t = 60$ and $\sigma = 3$. In the database shown in TABLE
\ref{tab:conversion}, item $\hat{p}_1$ is frequent. The $\hat{p}_1$-projected database
generated by PrefixSpan is:
\begin{align*}
(1) & ~d_1/\hat{p}_1 = \langle(\hat{p}_2,20),(\hat{p}_3,30)\rangle\\
(2) & ~d_2/\hat{p}_1 = \langle(\hat{p}_2,30),(\hat{p}_1,360),(\hat{p}_2,400),(\hat{p}_3,420)\rangle \\
(3) & ~d_4/\hat{p}_1 = \langle(\hat{p}_1,120), (\hat{p}_3,140), (\hat{p}_2,150), (\hat{p}_3,180)\rangle
\end{align*}
The elements satisfying $t \le 60$ are $(\hat{p}_2,20)$, $(\hat{p}_3, 30)$ and $(\hat{p}_2,30)$. No local item is frequent, hence
$\hat{p}_1$ cannot be grown any more.
\end{example}

To overcome this, given a sequence $\alpha$ and a frequent item
$\hat{p}$, we generate a postfix for every occurrence of $\hat{p}$.

\begin{example}
Also for Example 1, if we generate a postfix for every occurrence of
$\hat{p}_1$, the $\hat{p}_1$-projected database is:
\begin{align*}
(1) & ~d_1/\hat{p}_1 = \langle(\hat{p}_2,20),(\hat{p}_3,30)\rangle\\
(2) & ~d_2/\hat{p}_1 = \langle(\hat{p}_2,30),(\hat{p}_1,360),(\hat{p}_2,400),(\hat{p}_3,420)\rangle\\
(3) & ~d_2/\hat{p}_1 = \langle(\hat{p}_2,40),(\hat{p}_3,60)\rangle\\
(4) & ~d_4/\hat{p}_1 = \langle(\hat{p}_1,120), (\hat{p}_3,140), (\hat{p}_2,150), (\hat{p}_3,180)\rangle \\
(5) & ~d_4/\hat{p}_1 = \langle(\hat{p}_3,20), (\hat{p}_2,30), (\hat{p}_3,60)\rangle
\end{align*}
The items $\hat{p}_2$ and $\hat{p}_3$ are frequent and meanwhile satisfy the temporal constraint, thus longer patterns $\hat{p}_1
\ra{60} \hat{p}_2$ and $\hat{p}_1 \ra{60} \hat{p}_3$ are found in the projected database.
\end{example}




Using the above projection principle, the projected database includes all
postfixes to avoid missing patterns under the time constraint.  Algorithm
\ref{alg:prefixspan} sketches our algorithm for mining FEPs. The procedure is similar to the standard PrefixSpan algorithm in
\cite{PeiHPCDH01}, except that the aforementioned full projection principle is
adopted, and the time constraint $\Delta t$ is checked when searching for local
frequent items.



\begin{algorithm}
\label{alg:prefixspan}
\caption {Mining frequent evolving patterns.}
\small
\KwIn{support threshold $\sigma$, temporal constraint $\Delta t$, symbolic pollution database $\hat{P}$}
\SetKwProg{myproc}{Procedure}{}{}
\myproc{ InitialProjection($\hat{P}, ~\sigma, ~\Delta t$) }{
  $\L \gets$ frequent items in $\D$\;
  \ForEach{item $i$ in $\L$} {
    $S \gets \phi$\;
    \ForEach{sequence $o$ in $\hat{P}$} {
      $R \gets$ postfixes for all occurrences of $i$ in $o$\;
      $S \gets S \cup R$\;
    }
    PrefixSpan($i$, $i$, 1, $S$, $\Delta t$)\;		
  }  	
}
\SetKwProg{myfunc}{Function}{}{}
\myfunc{ PrefixSpan($\alpha$, $i_{prev}$, $l$, $S|_{\alpha}$, $\Delta t$) }{
  $\L \gets$ frequent items in $S|_{\alpha}$ meeting time constraint $\Delta t$\;
  \ForEach{item $i$ in $\L$} {
    \If{$i \ne i_{prev}$} {
      $\alpha' \gets$ append $i$ to $\alpha$\;
      Build $S|_{\alpha'}$ using full projection\;
      Output $\alpha'$\;
      PrefixSpan($\alpha'$, $i$, $l+1$, $S|_{\alpha'}$, $\Delta t$)\;
    }  	
  }
}
\end{algorithm}
\vspace{-3ex}
\begin{figure}[!htb]
  \centering
  \centerline{
    \includegraphics[width=1.0\columnwidth]{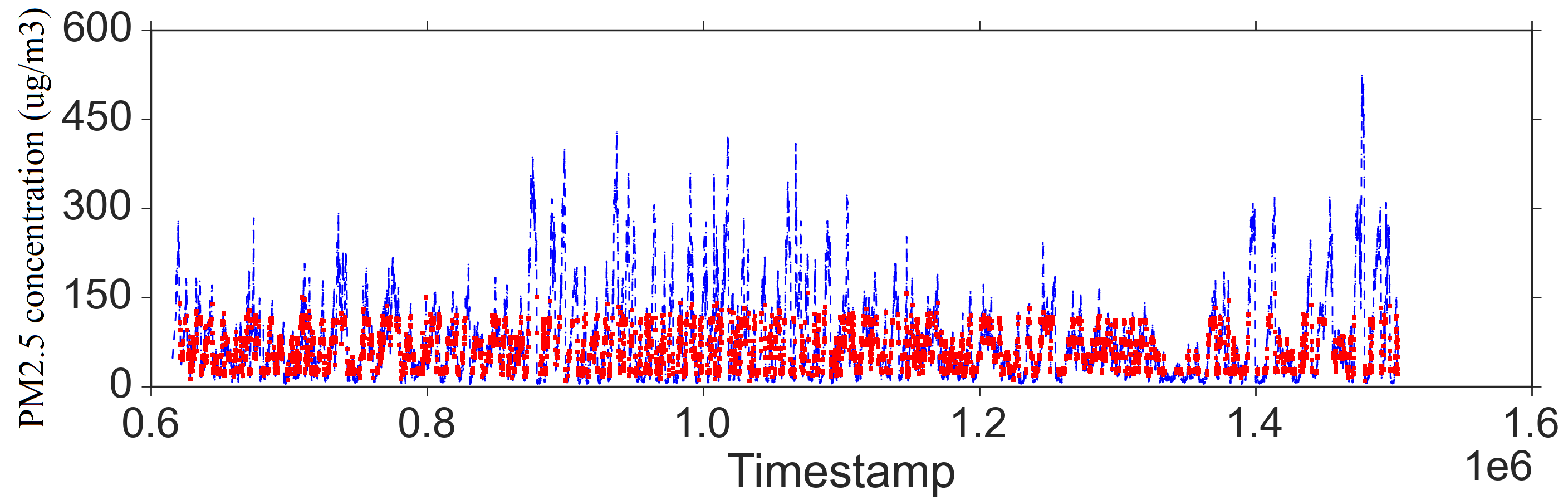}
  }
  
  \caption{An illustration of the pattern-matched timestamps. The blue dashed
    lines represents the PM2.5 time series in Beijing during a two-year period,
    and the red points denote the timestamps at which a certain FEP has
    occurred ($\sigma = 0.1$).}
  \label{fig:pattern-case}
\end{figure}

The output of Algorithm \ref{alg:prefixspan} is the set of all FEPs for the
given database, along with the occurring timestamps for each FEP. As an
example, Fig. \ref{fig:pattern-case} shows the raw PM2.5 time series in
Beijing during a two-year period. After mining FEPs on the symbolic pollution
database, we mark the timestamps at which the FEPs occur. One can observe that,
the FEPs can effectively capture the regularly appearing evolvements of PM2.5
in Beijing. Because of the support threshold and the evolving constraint,  
infrequent sudden changes and uninteresting fluctuations are all suppressed.

\subsection{Finding Candidate Causers}
\label{subsect:cand}

After discovering the FEPs, next step is leverage them to extract the candidate
causers for each sensor. Consider two sensors $s$ and $s'$, let us use $TS(s)$
and $TS(s')$ to denote the sets of pattern starting timestamps for $s$ and
$s'$, respectively.  Below, we introduce the \emph{pattern match} relationship.


\begin{definition}[Pattern Match]
  Let $t_{s'} \in TS(s')$ be a timestamp at which a pattern happens on $s'$.
  For a pattern starting timestamp $t_s \in TS(s)$, we say $t_{s'}$ matches
  $t_s$ if $0 \le t_{s} - t_{s'} \le L$, where $L$ is a
  pre-specified time lag threshold.
\end{definition}

Informally, the pattern match relation states that when there is a pattern
occurring on $s'$, then within some time interval, there is another pattern
happening on $s$. Naturally, if $s'$ has a strong causal effect on
$s$, then most timestamps in $TS_{s'}$ will be matched by $TS_s$, and vice
versa. Based on $TS_s$ and $TS_{s'}$, we proceed to introduce \emph{match
  precision} and \emph{match recall} to quantify the correlation
between $s$ and $s'$.

\begin{definition}[Match Precision]
  Given $TS_s$ and $TS_{s'}$, we define the matched timestamp set of $TS_{s'}$
  as $
  M_{s'} = \{t_{s'} | t_{s'} \in TS_{s'} \wedge \exists t_s \in TS_s,
  match(t_s, t_{s'}) = True\}.
  $
  With $M_{s'}$ and $TS_{s'}$, we define the precision of $s'$ matching $s$ as:
  $$ P(s, s') = |M_{s'}| / |TS_{s'}|$$
\end{definition}

\begin{definition}[Match Recall]
  Given $TS_s$ and $TS_{s'}$, we define the matched timestamp set of $TS_{s}$
  as
  $
  M_{s} = \{t_{s} | t_{s} \in TS_{s} \wedge \exists t_{s'} \in TS_{s'},
  match(t_s, t_{s'}) = True\}.
  $
  With $M_{s}$ and $TS_{s}$, we define the recall of $s'$ matching $s$ as:
  $$ R(s, s') = |M_{s}| / |TS_{s}|$$
\end{definition}

Relying on the concepts of \emph{match precision} and \emph{match recall}, we
compute the pattern-based correlation between $s$ and $s'$ as:
$$
Corr(s, s') = \frac{2 \times P (s, s')}{P(s, s') + R(s, s')}.
$$

Now we are ready to describe the process of finding candidate causers for each
sensor. Given the set of all sensors and their pattern-starting timestamps, our
goal is to find the candidate causers for each sensor.  Consider a target
sensor $s$, we say another sensor $s'$ is a candidate causer for $s$ if $s'$
satisfies two constraints: (1) the distance between $s$ and $s'$ is no larger
than a distance threshold $\delta_g$; and (2) the pattern correlation between
$s$ and $s'$ is no less than a correlation threshold $\delta_p$. Given the
pattern-starting timestamps that are ordered chronologically, the retrieval of
the candidate causers can be easily done by sequentially scanning the two
timestamp lists to find pattern-matched pairs.

Fig. \ref{fig:4seasons} illustrates eight examples of selected candidate causers. For PM2.5 in Beijing, we reduce the number of candidate sensors to $X=4 \sim 7$ from overall $|\S|=61$ sensors in North China. Note that China is a country with monsoon climate, the candidate sensors show quite similar geo-locations in four seasons. We therefore separate the training data into four groups based on seasons, to better diagnose causalities for the air pollutants in China.

\begin{figure}[!htb]
\label{fig:4seasons}

\centering
\includegraphics[width=3.1in]{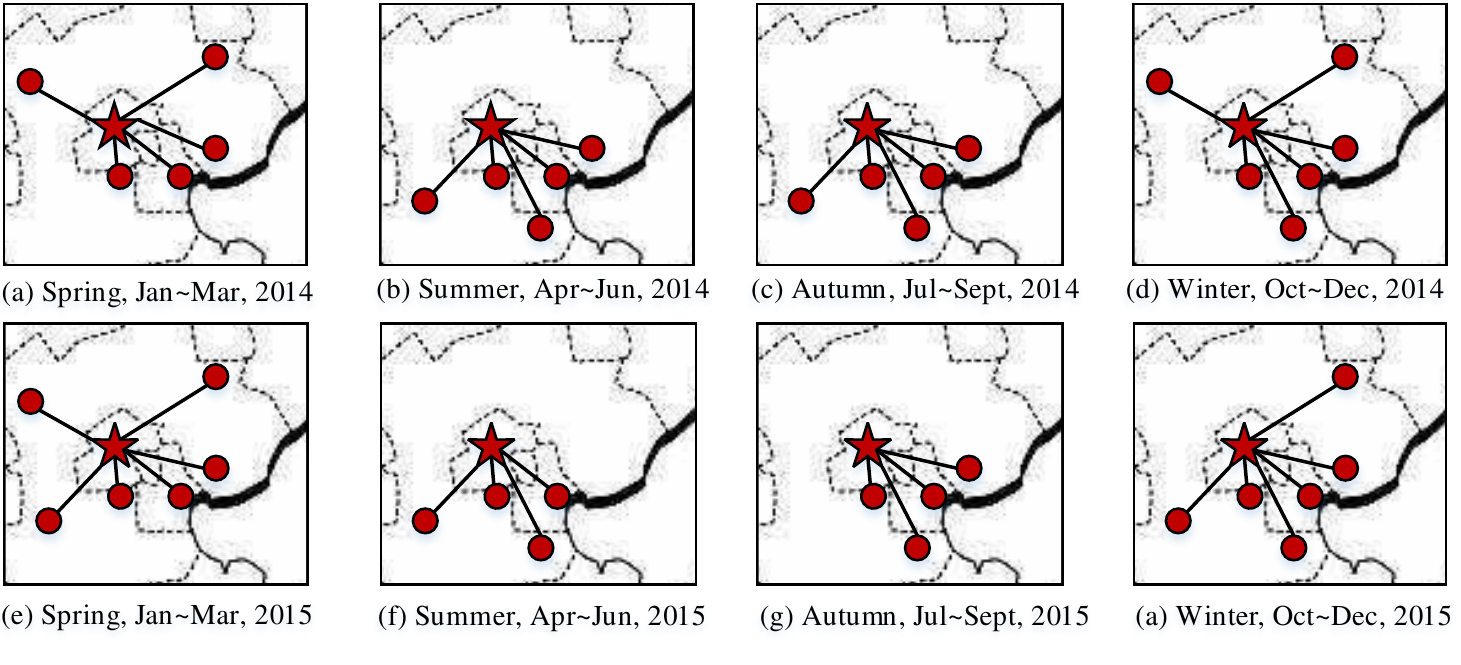}

\caption{\label{fig:4seasons} Candidate sensors for Beijing PM2.5 in four seasons. Star: PM2.5 in Beijing. Circles: pollutants at candidate sensors.}

\end{figure}

\section{The Bayesian Learning Module}
\label{sect:model}

In this section we first discuss how the causality learning benefits from the pattern-matched data extracted by the \emph{pattern mining module}. Then we dive into the methodology with the \emph{Bayesian learning module}. 

Identifying the ST causality (causal pathways) for air pollutants is a problem of learning the causal structures for multiple variables, which has been well discussed with the graphical causality \cite{pearl2003causality} based on Bayesian network (BN) \cite{heckerman1995learning}. Specifically, BN encodes the cause-and-effect relations in a directed acyclic graphs (DAG) via probabilistic dependencies. Learning BN structure from data is NP-complete \cite{chickering1996}, in the worst case requiring $2^{O(n^2)}$ searches among all the possible (DAGs). Thus when the number of variables becomes very large, the computational complexity will be unbearable. Therefore, we add the \emph{pattern mining module} before the \emph{Bayesian learning module} to combine the strengths of both. Pattern mining helps Bayesian learning by reducing the whole data to the selected candidate sensors and the periods matched by patterns, which greatly reduce the computational complexity as well as the noise in causality calculation. However, since the selected frequent patterns essentially demonstrates the ``correlation'', which is not ``causality'' \cite{holland}, the \emph{Bayesian learning module} helps represent and learn the causality.

Another benefit of conducting frequent pattern mining before Bayesian learning is that the selected frequent patterns could reflect the meaningful changes of the air pollutants, such as increase, decrease, sharp increase, sharp decrease, etc, thus significantly reducing the noises in Bayesian learning. When simply using Bayesian learning to identify the causality among different air pollutants time series, unreliable causal relations may be captured since there are many fluctuations and long-period low pollution cases which lead to unexpected correlation between two time series.

There are two major challenges to learn the causality among different pollutants in the ST space. The first one is to define a comprehensive representation of the causal pathways and diagnose the complex reactions and dispersions of different air pollutants. For example, the PM2.5 time series in Beijing can be strongly dependent on the NO2 time series locally, while it can also be influenced by the PM10 in another city. Therefore, both the local and ST dependencies need to be fairly considered in the model. We propose a Gaussian Bayesian network (GBN)-based graphical model, which captures the dependencies both locally and in the ST space. We elaborate how to generate initial causal pathways by GBN in Section \ref{sect:step3}. The second challenge is to learn faithful causal pathways given different weather conditions. As the example shown in Fig. 1, there could be different causal pathways under different wind speeds. We thus propose a method that integrates the meteorological data in the graphical model via a hidden factor representing the weather status (Section \ref{sect:step4}). In this way we can minimize the biases in the learning, and refine the final causal pathways (Section \ref{sect:step5}).

Here we give an example of combining the \emph{pattern mining module} with the \emph{Bayesian learning module}. Consider there are $|\S|$ monitoring sensors, with each sensor monitoring $M$ categories of pollutants, there will be $|\S| \times M$ variables in total for the Bayesian causal structure learning and the corresponding computational complexity will be $2^{O((|\S| \times M)^2)}$. When combining the \emph{pattern mining module}, we first extract the FEPs for each pollutant $P_{c_m s_n}$, i.e., the pollutant of category $m \in [1,M]$ collected at sensor $s_n \in \S$. Afterwards, for each target pollutant we select the pattern-matched periods (the timestamps that patterns at the neighborhood sensors happen ahead of the target sensor within some time interval, see Definition 6), as well as its top $|X|$ candidate causers (the $|X|$ neighborhood sensors that have the highest pattern-based correlation, see Definition 7 and 8). We then feed the pattern-matched periods selected and the candidate causers into the \emph{Bayesian learning module}. In this way the computational complexity is reduced to ${O(|X| \times M)}$, and the noises and fluctuations in the raw data are greatly suppressed.

\subsection{Generating Initial Causal Pathways}
\label{sect:step3}

This subsection first introduces the representation of causal pathways in the ST space, and then elaborates how to generate initial causal pathways.

\begin{definition} [Gaussian Bayesian Network (GBN)]
  \label{def:gbn}
GBN is a special form of Bayesian network for probabilistic inference with continuous Gaussian variables in a DAG, in which each variable is assumed as linear function of its parents \cite{GBNregression}.
\end{definition}

As shown in Fig. \ref{fig:gbn}, the ST causal relations of air pollutants are encoded in a GBN-based graphical model, to represent both local and ST dependencies. 
Here we choose GBN to model the causalities because: 1) GBN provides a simple way to represent the dependencies among multiple pollutants variables, both locally and in the ST space. 2) GBN models continuous variables rather than discrete values. Due to the sensors monitor the concentration of pollutants per hour, GBN could help better capture the fine-grained knowledge through the dependencies of these continuous values. In this subsection, based on the extracted matched patterns and candidate sensors from the \emph{pattern mining module} for each pollutant $\hat{P}_{c_m s_n}$, we use $P_{c_m s_n}$ to represent continuous values in the graphical model. 3) The characteristics of urban data fit the GBN model well. As shown in Fig. \ref{fig:distri}, the distribution of 1-hour difference (current value minus the value 1-hour ago) of air pollutants and meteorological data obey Gaussian distribution (verified by $D'Agostino-Pearson$ test \cite{d1973tests}\cite{zhu2015granger}). In the following sections, normalized 1-hour differences of time series data will be used as inputs for the model.

\begin{figure}[!htb]
\label{fig:gbn}

\centering
\includegraphics[width=2.5in]{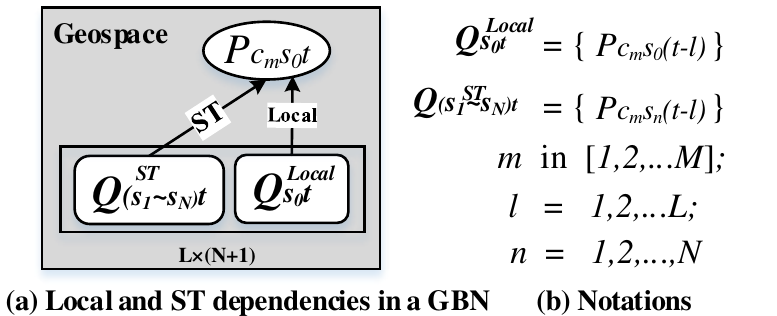}

\caption{\label{fig:gbn} GBN-based causal pathway representation and its notations.}

\end{figure}

\begin{figure}[!htb]
\label{fig:distri}

\centering
\includegraphics[width=3.2in]{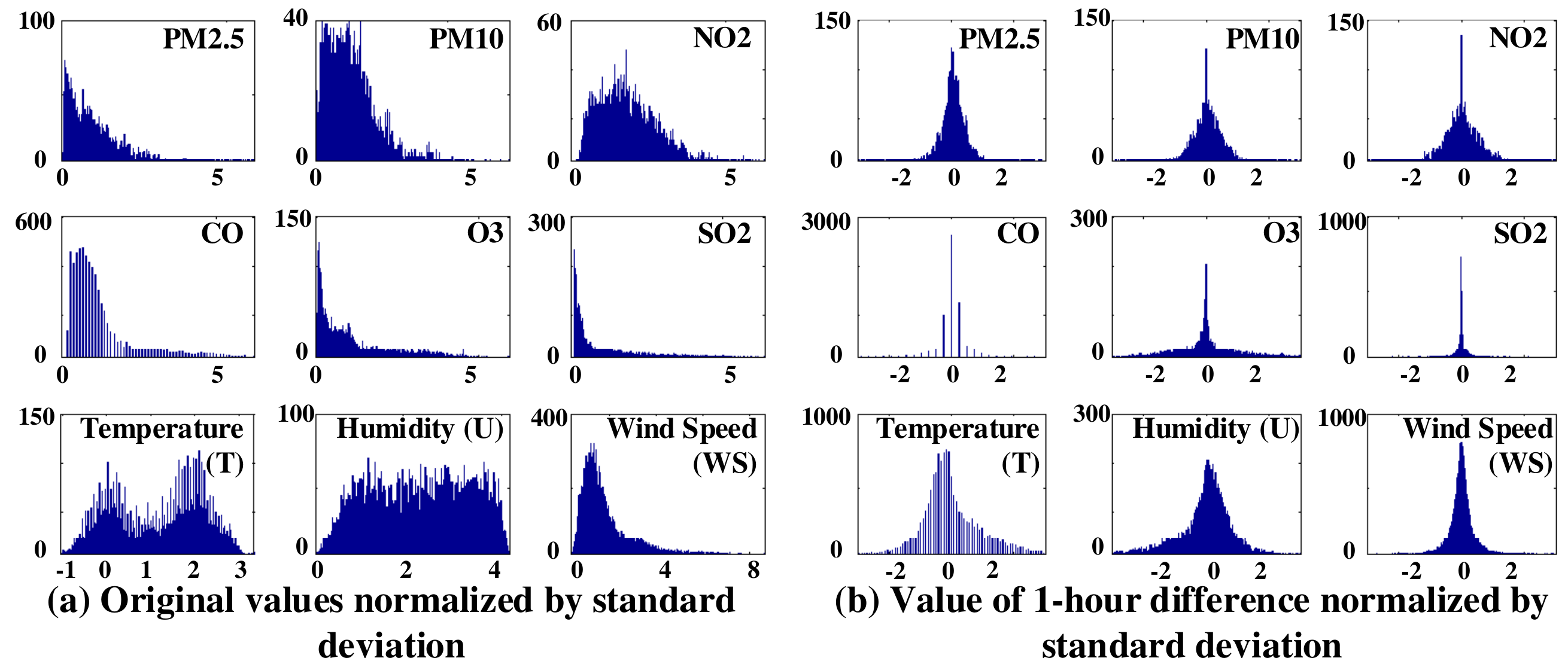}

\caption{\label{fig:distri} Histograms of urban data (original vs. 1-hour difference)}

\end{figure}

Specifically, for the target pollutant $c_m$ at sensor $s_0$-th sensor and timestamp $t$, denoted as $P_{c_m s_0 t},m \in[1,\mathrm{M}]$, we capture the dependencies from both the local causal pollutants $\bm{Q_{s_0 t}^{Local}}$ and the ST causal pollutants $\bm{Q_{(s_1 \sim s_N)t}^{ST}}$. Here $\bm{Q_{(s_1 \sim s_N)t}^{ST}}$ refer to a $1 \times \mathrm{NL}$ vector of pollutants at $\mathrm{N}$ neighborhood sensors $s_1 \sim s_N$ and previous $\mathrm{L}$ timestamps that most probably cause the target pollutant in the ST space, i.e. $\bm{Q_{(s_1 \sim s_N)t}^{ST}}=\{P_{c_{m_n} s_n (t-l) } \},m\in[1,\dots, \mathrm{M}]; n=1,\dots, \mathrm{N}; l=1,\dots, \mathrm{L}$. 
In order to better trace the most likely ``causers'' spatially, we just preserve the one category of pollutant at each neighborhood sensor that most influences the target pollutant. We use $c_{m_n}$ to represent the category for the most likely ``causers'' at sensor $n$.
Similarly, $\bm{Q_{s_0 t}^{Local}}$ is a $1 \times \mathrm{ML}$ vector of pollutants locally at $s_0$. For example, when we set $\mathrm{L}=2,\mathrm{M}=6$,  $\bm{Q_{s_0 t}^{Local}}$ may take values of 12 normalized 1-hour difference time series data, i.e. $\bm{Q_{s_0 t}^{Local}}=(2,-0.5,0.8,0.3,1,-2,2.2,1,1,0,-0.5,0.2)$.

The parents of $P_{c_m s_0 t}$ are denoted as $\bm{PA}(P_{c_m s_0 t})=\bm{Q_{s_0 t}^{Local}} \oplus \bm{Q_{(s_1 \sim s_N)t}^{ST}}$, where $\oplus$ denotes the concatenation operator for two vectors. Based on the definition of GBN, the distribution of $P_{c_m s_0 t}$ conditioned on $\bm{PA}(P_{c_m s_0 t})$ obeys Gaussian distribution:

\begin{equation}
\begin{aligned}
\label{eq:3}
&Pr⁡(P_{c_m s_0 t}=p_{c_m s_0 t} | \bm{PA}(P_{c_m s_0 t} ) ) \sim \mathcal{N}(\mu_{c_m s_0 t}+ \\
&\Sigma_{n=0}^\mathrm{N} \Sigma_{l=1}^\mathrm{L}  a_{{m_n}(n\mathrm{L}+l)}(p_{c_m s_n (t-l)}-\mu_{c_m s_n (t-l) } ),\Sigma(\epsilon_{c_m s_0 t} ))
\end{aligned}
\end{equation}

\noindent $\mu_{c_m s_0 t}$ is the marginal mean for $P_{c_m s_0 t}$. $\Sigma$ denotes the covariance operator. $\bm{A}=\{a_{m_n}(nL+l)\},(m_n\in[1,\dots,\mathrm{M}]; n=0,1,\dots,\mathrm{N}; l=1,\dots,\mathrm{L})$ is the coefficient for the linear regression in GBN \cite{GBNregression}:

To minimize the uncertainty of $P_{c_m s_0 t}$ given its parents, we need to find N sensors $s_1 \sim s_N$  from the ST space and the parameters $\bm{A}$ that minimize the error:
\begin{equation}
\begin{aligned}
\label{eq:5}
\Sigma(\epsilon_{c_m s_0 t})=\Sigma(P_{c_m s_0 t})-\bm{A} \Sigma(\bm{PA} (P_{c_m s_0 t}))^{-1} \bm{A}^T
\end{aligned}
\end{equation}

Generating the initial causal pathways requires locating N most influential sensors from $|\S|$ sensors with up to $\binom{|\S|}{N}$ trials. Yet given the candidate sensors selected by Section \ref{subsect:cand}, we manage to search from a subset $(X \le |\S|)$ sensors with time efficiency and scalability. 
We further propose a Granger causality score $GC_{score}$ to generate initial causal pathways, which is defined as:

\begin{equation}
\begin{split}
GC_{score} (m,s_0,s_n )&=max_{m_n\in [1,\mathrm{M}]}max_{l\in [1,\mathrm{L}]}\\\{|match(t_{(c_m,s_0)}, t_{(c_{m_n},s_n)})| 
&\cdot \frac{|\Sigma(\epsilon_{c_m s_0 (t-l)})_1|-|\Sigma(\epsilon_{c_m s_0 (t-l)})_2|}{|\Sigma(\epsilon_{c_m s_0 (t-l)})_2|∙\chi_\mathrm{L}^2(0.05)}\}
\end{split}
\end{equation}

\noindent where $GC_{score}$ is a $\chi^2$-test score \cite{hlavavckova2007causality} for the predictive causality, with higher score indicating more 
probable ``Granger'' causes from M pollutants at sensor $s_n$ to the target pollutant $c_m$ at sensor $s_0$ \cite{granger1969investigating} 
($GC_{score} \le 1$ means none causality). For variables obeying Gaussian distribution, Granger causality is in the same form as conditional mutual information \cite{barnett2009granger}, which has been used successfully for constructing structures for Bayesian networks. Here $|match(t_{(c_m,s_0)}, t_{(c_{m_n},s_n)})|$ is the number of matched timestamps of FEPs between two time series (pollutant $c_{m_n}$ at sensor $s_n$ and pollutant $c_m$ at sensor $s_0$, see Section \ref{subsect:cand}). And $\Sigma(\epsilon_{c_m s_0 (t-l)})_1$ and $\Sigma(\epsilon_{c_m s_0 (t-l)})_2$ correspond to the variances of the target pollutant $P_{c_m s_0 t}$ conditioned on lagged sequences $\bm{Q_{s_0 (t-l)}^{Local}}$ and $\bm{Q_{s_0 (t-l)}^{Local}} \oplus \bm{Q_{s_n (t-l)}^{ST}}$.






\subsection{Integrating Confounders}
\label{sect:step4}

Recall the example in Fig. \ref{fig:example}. A target pollutant is likely to have several different causal pathways under different environmental conditions, which indicate the causal pathways we learn may be biased and may not reflect the real reactions or propagations of pollutants. To overcome this, it is necessary to model the environmental factors (humidity, wind, etc.) as extraneous variables in the causality model, which simultaneously influence the cause and effect. For example, when the wind speed is less than 5m/s, city A's PM2.5 could be the ``cause'' of city B's PM10. However, when the wind speed is more than 5m/s, there may not be causal relations between the two pollutants in the two cities. In this subsection, we will elaborate how to integrate the environmental factors into the GBN-based graphical model, to minimize the biases in causality analysis and guarantee the causal pathways are faithful for the government's decision making. We first introduce the definition of confounder and then elaborate the integration.

\begin{definition} [Confounder]
  \label{def:confounder}
A confounder is defined as a third variable that simultaneously correlates with the cause and effect, e.g. gender $K$ may affect the effect of recovery $P$ given a medicine $Q$, as shown in Fig. \ref{fig:integrating}(a). Ignoring the confounders will lead to biased causality analysis. To guarantee an unbiased causal inference, the cause-and-effect is usually adjusted by averaging all the sub-classification cases of $K$ \cite{pearl2003causality}, i.e. $Pr⁡(P | do(Q) )=\Sigma_{k=1}^K Pr⁡(P|Q,k)Pr⁡(k)$. 
\end{definition} 

\begin{figure}[!htb]
\label{fig:integrating}

\centering
\includegraphics[width=3.3in]{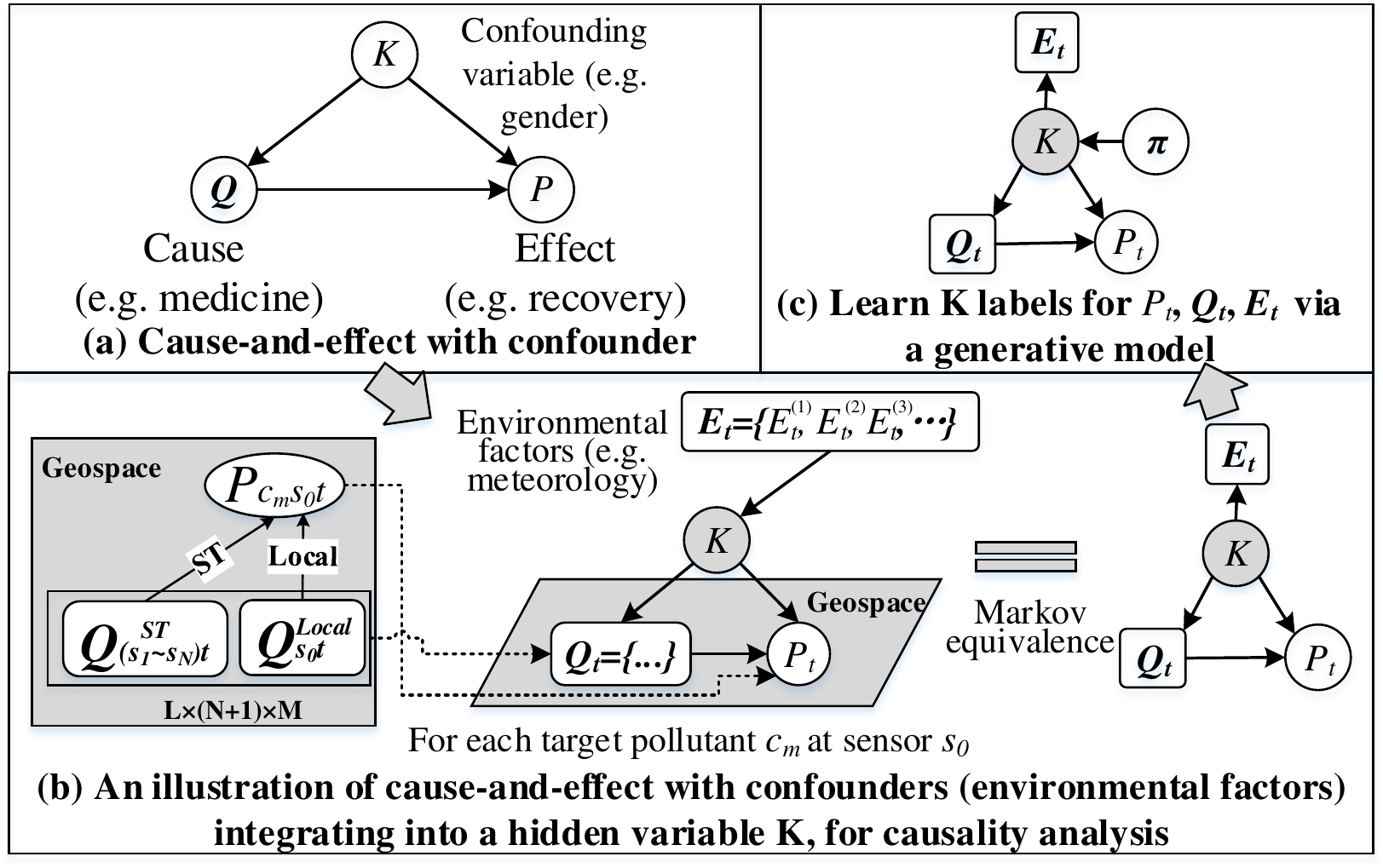}

\caption{\label{fig:integrating} The GBN-based graphical model, integrating confounders to the causal pathway, and converting the model into a generative model}

\end{figure}

For integrating environmental factors as confounders, denoted as $\bm{E_t}=\{E_t^{(1)},E_t^{(2)},\dots\}$, into the GBN-based causal pathways, one challenge is there can be too many sub-classifications of environmental statuses. For example, if there are 5 environmental factors and each factor has 4 statuses, there will exist $4^5=1024$ causal pathways for each sub-classification case. Directly integrating $\bm{E_t}$ as confounders to the cause and effect will result in unreliable causality analysis due to very few sample data conditioned on each sub-classification case. Therefore, we introduce a discrete hidden confounding variable $K$, which determines the probabilities of different causal pathways from $\bm{Q_t}$ to $\bm{P_t}$, as shown in Fig. \ref{fig:integrating}(b). The environmental factors $\bm{E_t}$ are further integrated into $K$, where $K = 1,2,...\mathrm{K}$. In this ways, the large number of sub-classification cases of confounders will be greatly reduced to a small number $\mathrm{K}$, as K clusters of the environmental factors.

Based on Markov equivalence (DAGs which share the same joint probability distribution \cite{flesch2007markov}), we can reverse the arrow $\bm{E_t} \to K$ to $K \to \bm{E_t}$, as shown in the right part of Fig. \ref{fig:integrating}(b). $K$ determines the distributions of $P,\bm{Q_t,E_t}$, thus enabling us to learn the distribution of the graphical model from a generative process. To help us learn the hidden variable $K$, the generative process further introduces a hyper-parameter $\bm{\pi}$ (as shown in Fig. \ref{fig:integrating}(c)) that determines the distribution of $K$. Thus the graphical model can be understood as a mixture model under K clusters. We learn the parameters of the graphical model by maximizing the new log likelihood:




\begin{equation}
\label{eq:ll}
\begin{split}
LL^{gen}=\Sigma_t \Sigma_{k=1}^\mathrm{K} ln⁡(Pr(p_t | \bm{q_t},k)Pr⁡(\bm{e_t} | k)Pr⁡(k|\bm{\pi})) 
\end{split}
\end{equation}

In determining the number of the hidden variable $K$, we do not consider too large K values since that will induce much complexity for causality analysis. Also a too small K may not characterize the information contained in the confounders (i.e. meteorology). We observe the 2-D PCA projections of meteorological data (as shown in Fig. \ref{fig:weatherclusters}). In three regions, five clusters can characterize the data sufficiently well. Thus we choose $\mathrm{K} = 3 \sim 7$ for learning in practice.

 \begin{figure}[!htb]
\label{fig:weatherclusters}

\centering
\includegraphics[width=3.0in]{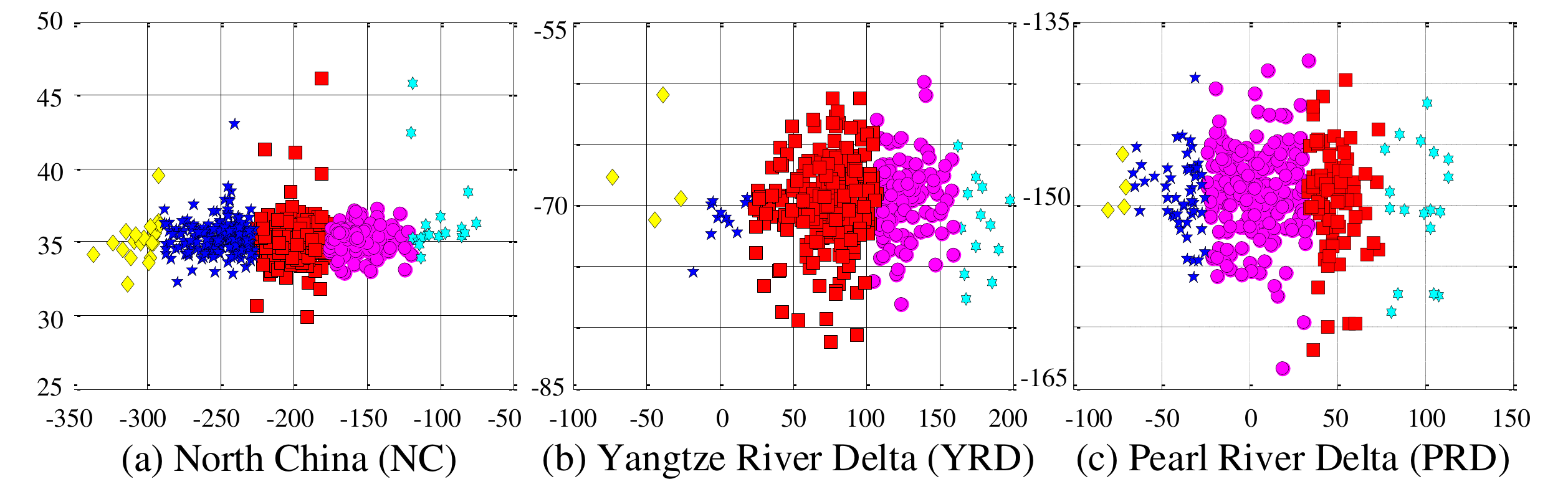}

\caption{\label{fig:weatherclusters} 2-D PCA projections of 5 clusters of meteorological data in NC, YRD and PRD. The original meteorological data contains five types, i.e. temperature (T), pressure (P), humidity (U), wind speed (WS), and wind direction (WD), with each region divided into 9 grids, thus 45-dimensional.}

\end{figure}

\subsection{Refining Causal Structures}
\label{sect:step5}

This subsection tries to refine the causal structures and obtain the final causal structures under K clusters. The refining process includes two phases in each iteration: 1) an EM learning (EML) phase to infer the parameters of the model, and 2) a structure reconstruction (SR) phase to re-select the top N neighborhood sensors based on the newly learnt parameters and $GC_{score}$, as illustrated in Algorithm \ref{alg:refine}.

EML (line 6-18) is an approximation method to learn the parameters $\bm{\pi, \gamma, A_k,B_k}$ of the graphical model, by maximizing the log likelihood (Equation \ref{eq:ll}) of the observed data sets via an $E$-step and a $M$-step. Here $\bm{\pi}$ contains the hyper parameters which determine the distribution of K ($\mathrm{T} \times \mathrm{K}$-dimensional). $\bm{\gamma}$ are posterior probabilities for each monitoring record ($\mathrm{T} \times \mathrm{K}$-dimensional). $\bm{A_k,B_k}$ are parameters for measuring the dependencies among pollutants and meteorology ($\mathrm{K}$-dimensional). Note that $\bm{A_k,B_k}$ come in different formats. $\bm{A_k}$ is the regression parameter for:

\begin{equation}
\label{eq:regression}
P_{c_m s_0 t}=\mu_0+
(\bm{Q_{s_0 t}^{Local}} \oplus \bm{Q_{(s_1 \sim s_N)t}^{ST}})\bm{A_k} +\epsilon_{c_m s_0 t}
\end{equation}
and $\bm{B_k}=(\bm{\mu_{B_k},\Sigma_{B_k}})=(mean(\bm{E_t}), std(\bm{E_t}))$ includes the parameters for the multivariate Gaussian distribution of environmental factors $\bm{E_t}$. In the $E$-step, we calculate the expectation of log likelihood (Equation \ref{eq:estep}) with the current parameters, and the $M$-step re-computes the parameters.

\pa{0ex}{$\bm{E}$-step:}
Given the parameters $\bm{\pi}, \mathrm{K,N}, \bm{A_k,B_k}$, EM assumes the membership probability $\gamma_{tk}$, i.e., the probability of $p_t,\bm{q_t,e_t}$ belonging to the $k$-th cluster as:

\begin{equation}
\label{eq:estep}
\begin{split}
\gamma_{tk}&=Pr⁡(k|p_t,\bm{q_t,e_t})=\dfrac{Pr⁡(k)Pr⁡(p_t,\bm{q_t,e_t}|k)}{Pr⁡(p_t,\bm{q_t,e_t})}\\
&=\dfrac{\pi_{tk} \mathcal{N}(p_t |\bm{q_t},\bm{A_k})\mathcal{N}(\bm{e_t}|\bm{B_k})}{\Sigma_{j=1}^K \pi_{tj} \mathcal{N}(p_t |\bm{q_t},\bm{A_j})\mathcal{N}(\bm{e_t}|\bm{B_j})}
\end{split}
\end{equation}

\pa{0ex}{$\bm{M}$-step:}
The membership probability $\gamma_{tk}$ in $E$-step can be used to calculate new parameter values $\bm{\pi^{new}}, \bm{A_k^{new},B_k^{new}}$. We first determine the most likely assignment tag of timestamp $t$ to cluster $k$, i.e.

\begin{equation}
\label{eq:labelt}
\begin{split}
Tag_t=max_{k \in [1,\mathrm{K}] }⁡\pi_{tk}
\end{split}
\end{equation}

By integrating the timestamps belonging to each cluster $k$, we can update $\bm{A_k^{new}}$ by Equation \ref{eq:regression}. Then we update $\bm{B_k}$ by:

\begin{equation}
\label{eq:bk}
\begin{split}
\bm{\mu_{B_k}^{new}}&=\dfrac{1}{T_k}  \Sigma_{t=1}^T \gamma_{tk}\bm{e_t}, T_k=\Sigma_{t=1}^T \gamma_{tk} \\
\bm{\Sigma_{B_k}^{new}}&=\dfrac{1}{T_k} \Sigma_{t=1}^T \gamma_{tk}(\bm{e_t}-\bm{\mu_{B_k}^{new}})(\bm{e_t}-\bm{\mu_{B_k}^{new}})^T
\end{split}
\end{equation}

In addition, we update $\pi_{tk}^{new}$ by:

\begin{equation}
\label{eq:pi}
\begin{split}
\pi_{tk}^{new}=\dfrac{\gamma_{tk}}{T_k}
\end{split}
\end{equation}

The SR phase (line 19-24) utilizes the parameters provided by the EM learning phase, and re-select the top N neighborhood sensors based on the newly generated $GC_{score}$ for each cluster $k$. 
We present a training example (as shown in Fig. \ref{fig:training}(a)) of learning the causal pathways for Beijing PM2.5 during Jan$-$Mar. After 20 training iterations of the EM learning phase and structure reconstruction, 
we finally obtain $\mathrm{K}=4$ causal structures under each cluster, with the log likelihood shown in Fig. \ref{fig:training}(b). We find the log likelihood does not increase much after 10 iterations, thus we set the iteration number to 10 in our experiments.
For the last iteration, we calculate the percentage of labeled timestamps belonging to each cluster $k$. In this example, we find that Beijing's PM2.5 is more likely to be influenced by NO2 in Baoding and PM10 in Cangzhou.

\begin{algorithm}[!htb]
\label{alg:refine}
\small
\caption {Refining the causal structures for each target pollutant $c_m$ at location $s_0$.}
\KwIn{$\mathrm{T,K,N}$, and raining data sets $p_t$, $\bm{q_t}$, $\bm{e_t}$, $t \in [1,\mathrm{T}]$}
\KwOut{Refined causal structures for K clusters}
\SetKwProg{myproc}{Procedure}{}{}
Initial neighborhood sensors $s_1 \sim s_N$ based on top $N$ $GC_{score}$;\\
\Repeat{ Log\_likeoihood converges }{
  EML($P_t$, $\bm{Q_t}$, $\bm{E_t}$, $s_1 \sim s_N$, $\mathrm{K}$) $\to Log\_likelihood, \pi_{tk},\gamma_{tk}, \bm{A_k,B_k}$;\\
  SR($\bm{A_k}$, $s_1 \sim s_N$, $\mathrm{K}$)$\to s'_1 \sim s'_N$, $Q'$;
}
\SetKwProg{myfunc}{Function}{}{}
\myfunc{ EM\_Learning(EML)($P_t$, $\bm{Q_t}$, $\bm{E_t}$, $s_1 \sim s_N$, $\mathrm{K}$)}{
  \Repeat{ Log likelihood converges }{
  InitialAssign: $K$ clusters via K-means($\bm{E_t}$)\\
  \ForEach{item $t = 1$ to $\mathrm{T}$} {
    \ForEach{item $k = 1$ to $\mathrm{K}$} {
      Update $\pi_{tk}$ by Equation (\ref{eq:pi});
    }		
  }
  \ForEach{item $k = 1$ to $\mathrm{K}$} {
    Update $\bm{A_k,B_k}$ by Equation (\ref{eq:regression}),(\ref{eq:bk});
  }		
  \ForEach{item $t = 1$ to $\mathrm{T}$} {
    \ForEach{item $k = 1$ to $\mathrm{K}$} {
      Update $\gamma_{tk}$\ by Equation (\ref{eq:estep});
    }		
  }
}
return: $Log\_likelihood$ and $\pi_{tk},\gamma_{tk}, \bm{A_k,B_k}$;
}
\SetKwProg{myfunc}{Function}{}{}
\myfunc{ Structure\_Reconstruction(SR)($\bm{A_k}$, $s_1 \sim s_N$, $\mathrm{K}$)}{
  \ForEach{item $s_n$ in All candidate sensors} {
    Compute $GC_{score}(m,s_0,s_n)$ for $s_1 \sim s_N$;\\
	Rank $GC_{score}$ and re-select the top N neighborhood sensors $s'_1 \sim s'_N$;\\
	Update $Q \to Q'$ corresponding to $s'_1 \sim s'_N$;
  }		
return: $s'_1 \sim s'_N$, $Q'$;
}
\end{algorithm}

\begin{figure*}[!htb]
\label{fig:training}

\centering
\includegraphics[width=7in]{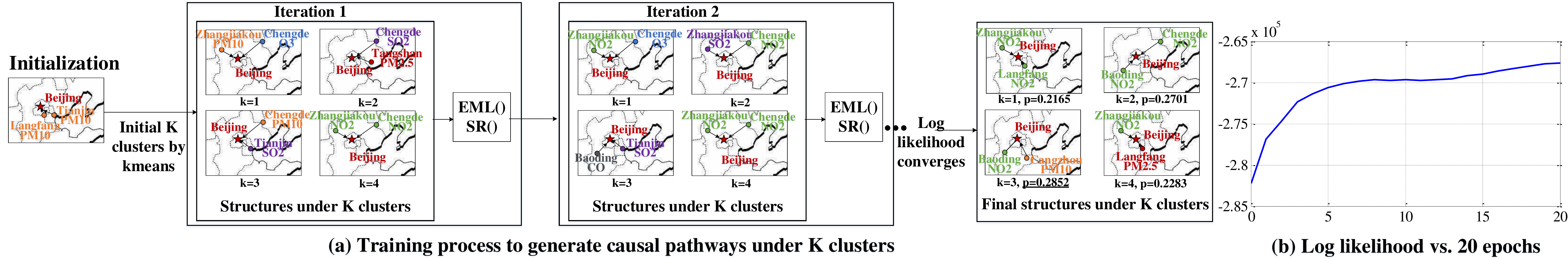}

\caption{\label{fig:training} An example of learning the causal pathway for PM2.5, Jan$-$Mar in Beijing under $K=4$ clusters.}

\end{figure*}

\section{Experiments}
\label{sect:exp}

We evaluate the empirical performance of our method in this section. All the
experiments were conducted on a computer with Intel Core i5 3.3Ghz CPU and 16GB
memory. We use MATLAB for our Bayesian learning module, and the open-source
MATLAB BNT toolbox~\cite{murphy2001bayes} for baseline methods.

\subsection{Experimental Setup}
\label{sect:setup}

\subsubsection{Data Sets} We use three data sets that contain the records of 6 air pollutants and 5 meteorological measurements:

$\bullet$ North China (NC), with 61 cities, 544 air quality monitoring
sensors and 404 meteorological sensors in North China.
The latitude and longitude ranges are 34N-43N,
110E-123E. 

$\bullet$ Yangtze River Delta (YRD), with 49 cities, 330 air quality
monitoring sensors and 48 meteorology sensors. The latitude and longitude ranges are
28N-35N, 115E-123E, respectively. 

$\bullet$ Pearl River Delta (PRD), with 18 cities, 124 air quality
monitoring sensors and 406 meteorology sensors. The latitude and longitude
ranges are 22N-25N, 110E-116E. 

The 6 air pollutants are PM2.5, PM10, NO$_2$, CO, O$_3$, SO$_2$, and the 5 meteorological
measurements are temperature (T), pressure (P), humidity (H), wind speed (WS),
and wind direction (WD), which are updated hourly. The time span for all data sets is from
01/06/2013 to 31/12/2016. We separate each data set into four groups based on
four seasons, and use the last 15 days in each season in year 2014, 2015, 2016 for
testing, and the remaining data for model training. The total numbers of training timestamps are 5424, 6193, 7753, 7752 in the four seasons, respectively, 
and the number of the corresponding testing timestamps is 15$\times$24$\times$3=1080 in each season.
To get 
the environmental factors $\bm{E_t}$ for
the coupled model, we divide each region into $3\times3$ grids and average the meteorology values within
each grid.

We conduct experiments at both city level (Section \ref{sect:time}, \ref{sect:accuracy}, \ref{sect:casestudy}) and
sensor level (Section \ref{sect:scalability}).  The city-level experiments average value of the
sensors in the city to form a pseudo sensor, and discover the pathways among
all the cities in three data sets. The sensor-level experiments analyze
the causal relationships among sensors in each data set.

\subsubsection{Baselines} Since Bayesian-based methods have been well used to learn causal Bayesian structures \cite{heckerman1995learning}, we choose the most commonly used BN
structure learning approaches as baselines to compare with our method. To identify the dependencies among different pollutants,
the baselines are deployed to learn the causal structures for each target pollutant.

\pa{0ex}{1. MWST.} Maximum Weighted Spanning Tree (MWST) generates an
undirected tree structure based on the MWST algorithm~\cite{chow1968mwst}. Each
time it connects one edge between two nodes with the maximum mutual
information. Furthermore, \cite{polytree} proposed an independency
test method to assign a direction to each edge in the tree structure.

\pa{0ex}{2. MCMC.} Markov-chain Monte Carlo (MCMC) is a statistical method that
also samples from the Directed Acyclic Graph (DAG) space \cite{beck2002mcmc}.
The method maximizes the score from a set of similar DAGs that add,
delete, or reverse connections, and updates the structure in the next
iteration. 

\pa{0ex}{3. K2+PS.} $K2$ is a widely used greedy method for Bayesian
structure learning, which selects at most N parents based on the $K2$ score
\cite{cooper1992bayesian} for each variable given the updating order of all the
variables.  In our case, we use pattern search
algorithm~\cite{lewis2002patternsearch} to optimize the updating order, thus
reducing the search space of casual pathways of different pollutants.  Note
that the original $K2$ score is defined for discrete variables. Here we use
$GC_{score}$ instead for the continuous variables.

\pa{0ex}{4. CGBN.} Coupled Gaussian Bayesian network \cite{zhu2016gcm} is a data-driven causality model considering the dependencies between both the air pollutants and meteorology. 
CGBN assumes there is a third variable (confounder, such as gender as a confounder to evaluate the effect of a medicine on a disease) which 
simultaneously influences the dependences among pollutants and among environmental factors, coupling pollutants and environmental factors together. The difference between CGBN and our approach
is that 1) our approach integrates the environmental factors directly into the graphical model, instead of through coupling, and 2) our approach has a pattern mining module and a refining algorithm to optimize the learning process.

\subsubsection{Parameter Setting}

The parameters of \emph{pg-Causality} include: 
(1) the support threshold $\sigma$;
(2) the temporal constraint $\Delta t$; (3) the distance threshold $\delta_g$
for finding candidate causers; and (4) the correlation threshold $\delta_p$ for
finding candidate causers; (5) the number of time lags L = 3; (6) and the number of pollutant categories M = 6. When finding causal pathways at city level, we set
$\sigma = 0.1$, $\Delta t = 1$ hour, $\delta_g = 200$ km, and $\delta_p = 0.5$.
At the station level, all the the parameters are set the same except that
$\delta_g = 15$ km to impose a finger granularity for finding candidate causers. K and N are evaluated within the range K $= 3 \sim 7$, and N$=1 \sim 5$.


\subsection{Experimental Results}
\label{sect:results}
The verification of causality is a very critical part in causal modelling. The simplest method for evaluating causal dependence is to intervene in a system and determine if the model is accurate under intervention. However, 
substantial and direct intervention in air pollution is impossible. By investigating the verification methods in previous causality works, we propose five tasks to
evaluate the effectiveness of our approach, namely, 1) inference accuracy for a 1-hour prediction task, 2) time efficiency, 3) scalability, 4) verification on synthetic data, and 5) visualizing the causal pathways.
Tasks 1-3 target to evaluate whether the model fits the dependences among the datasets well. Task 4 tries to learn the causal pathways for a predefined causal structure generated by synthetic datasets. And Task 5 targets at the 
interpretability of the causal pathways we learn.

\subsubsection{Inference Accuracy}
\label{sect:accuracy}
We first evaluate the effectiveness of our approach via the causal inference accuracy through the causal pathways at city level, which is a 1-hour prediction task based on our proposed GBN-based graphical model. Note this prediction task is not general for all the timestamps, it only predicts the future 1-hour based on the extracted pattern-matched periods, indicating the causal inference for the frequent evolving behaviors.
Specifically, we first infer the probability $Pr(k)$ of the testing data belonging to cluster $k$. Then, we use the structure and parameters from the trained causal pathways regarding this cluster to estimate the future pollutant concentration by Eq.~\ref{eq:predict}.
\begin{equation}
\label{eq:predict}
\begin{split}
P_{c_m s_0 t}^{est}=\Sigma_{k=1}^\mathrm{K}(\mu_{0k}+\bm{PA}(P_{c_m s_0 t})\bm{A_k})Pr(k)
\end{split}
\end{equation}

The accuracy is defined as $\Sigma_{t=1}^\mathrm{T_{test}}( P_{c_m s_0 t}^{est} - P_{c_m s_0 t}^{*} ) / P_{c_m s_0 t}^{*}\mathrm{T_{test}} $, where $P_{c_m s_0 t}^{*}$ is the ground truth value and $\mathrm{T_{test}} $ is the number of test cases. TABLE \ref{table:accuracy} shows the 1-hour prediction accuracy for PM2.5 and PM10 with our approaches \emph{pg-Causality}, \emph{pg-Causality-n}, \emph{pg-Causality-p}, and the three baseline methods in Beijing (Region NC), Shanghai (Region YRD), and Shenzhen (Region PRD). Here \emph{pg-Causality-n} represents \emph{pg-Causality} without the pattern mining module, and \emph{pg-Causality-p} represents \emph{pg-Causality} without integrating confounders.
The accuracy shown in TABLE~\ref{table:accuracy} is the accuracy for spring for three cities. The \emph{pg-Causality} gets the highest accuracy (92.5\%, 93.78\%, 95.39\% for PM2.5 in Beijing, Shanghai, and Shenzhen, respectively; 91.36\%, 92.39\%, 93.18\% for PM10, repectively.), compared to \emph{pg-Causality-n} and \emph{pg-Causality-p}, as well as the three baseline methods WMST, K2+PS, and CGBN. We did not include the accuracy of MCMC in TABLE \ref{table:accuracy} due to its unbearably high computational time. The accuracy for MCMC is lower than 60\%, which is not competitive with the other methods mentioned. The highest inference accuracy for the three cities are marked with three different colors (orange for Beijing, blue for Shanghai, and green for Shenzhen) given different parameters $\mathrm{K}$ and $\mathrm{N}$. $\mathrm{K}$ and $\mathrm{N}$ are obtained based on the maximum inference accuracy for each city. We note $\mathrm{N}=2,\mathrm{K}=4$ provides the best performance for Beijing, while $\mathrm{N}=0,\mathrm{K}=5$ or $6$ generate the best accuracy for Shanghai and $\mathrm{N}=0,\mathrm{K}=1$ for Shenzhen. The optimal number $\mathrm{N}=2$ for Beijing also suggests that the air pollution is mainly influenced by the most influential sensors in the ST space. While the optimal number $\mathrm{N}=0$ for Shanghai and Shenzhen suggests that the PM2.5 in these two cities are mainly influenced by historical pollutants locally. 

We also
evaluate the 1-hour prediction accuracy with three well-used time
series model, i.e., auto-regression moving average (ARMA) model,
linear regression model (LR), and support vector machine for regression with a Gaussian radial basis function (rbf) kernel (represented as SVM-R). Generally, \emph{pg-Causality} demonstrates higher inference accuracy compared with these time series models, except for the PM2.5 in Shanghai.

\begin{table*}[!htb]
\label{table:accuracy}
\caption{\label{table:accuracy}  Accuracy of PM2.5/PM10 1-hour prediction vs. baselines, Beijing, Shanghai, and Shenzhen.}

\centering
\includegraphics[width=7in]{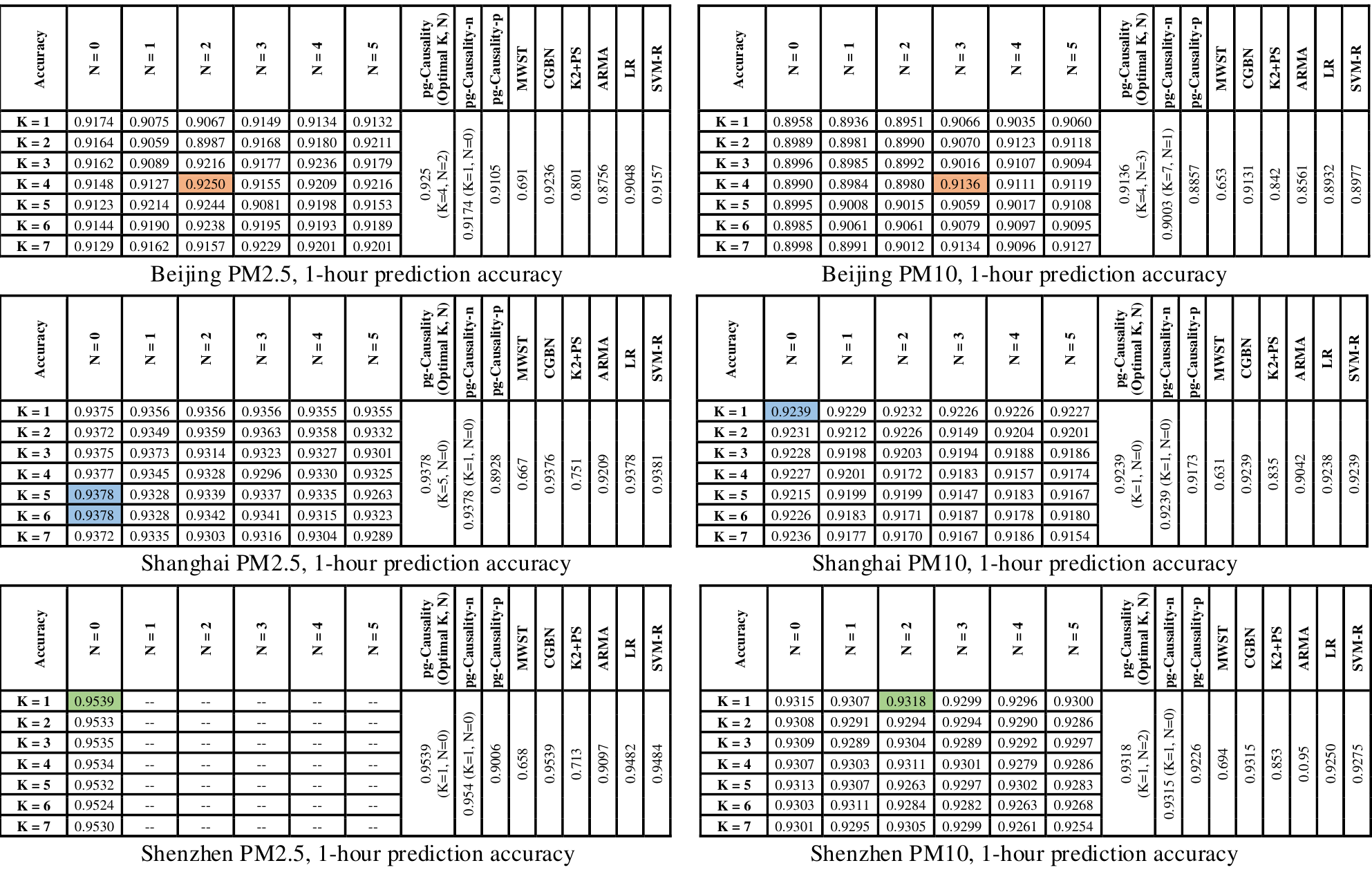}

\end{table*}

\subsubsection{Time efficiency}
\label{sect:time}
We also compare the training time of \emph{pg-Causality} with baseline methods, as shown in TABLE \ref{table:time}. Since our approach consists of both pattern mining and Bayesian learning modules, we present the averaged time consumption of training all the three data sets, for each step in the two modules. We also evaluate the overall time consumption of \emph{pg-Causality} and \emph{pg-Causality-n} without the pattern mining module (Section 5.1 (p+g) refers to the time cost of causal structure initialization with both pattern mining and Granger causality score. Section 5.1 (g) refers to only using Granger causality score). Results show that our approach is very efficient, with the second minimum computation time among all the methods. MWST consumes the minimal time, however, it does not generate satisfactory accuracy for prediction (as in Section \ref{sect:accuracy}). We thus consider our approach provides the best trade-off regarding accuracy and time efficiency.

\begin{table}[!htb]

\label{table:time}
\caption{\label{table:time} Computation time for training data sets at city level.} 

\centering
\includegraphics[width=3.4in]{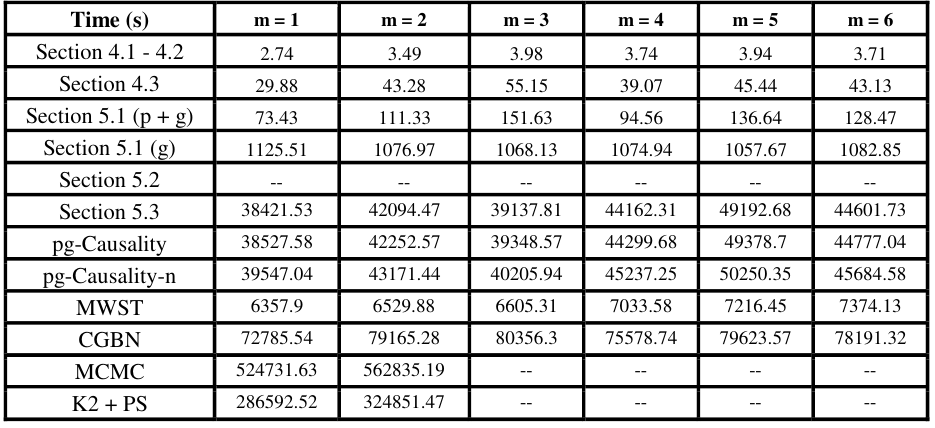}

\end{table}

\subsubsection{Scalability}
\label{sect:scalability}
Another superior characteristic of our approach is the scalability. We further identify the causal pathways for air pollutants at sensor level, which is more than ten times as large as in the city-level analysis. Our approach provides linear scalability in time with 11.6 hours training time at city level for 128 cities, and 126 hours at sensor level for 982 stations.
We here claim linear scalability since we did not try to find the optimal causal structure by searching the DAG space, which is an NP hard problem and in the worst case requires $2^{O(n^2)}$ searches \cite{chickering1996}.
In this paper, the causal pathways we learnt are based on greedy-based approximation. For the structure learning algorithm, we assume the number of parameters of the Bayesian-based graphical model to be (\#), and the training iterations to be $N_{iter}$. For totally $\mathrm{N}$ sensors in the geospace and $\mathrm{T}$ timestamps in the training records, the time cost for the EM learning (EML) phase is $O(N_{iter}\times(\#)\times\mathrm{N}\times\mathrm{T})$, assuming every parameter is updated once for every record. In addition, the time cost for the structure reconstruction (SR) phase is $O(N_{iter}\times\mathrm{X}\times\mathrm{L}\times\mathrm{N}\times\mathrm{T} + N_{iter}\times\mathrm{K}\times(\#)\times\mathrm{N}\times\mathrm{T})$, where $\mathrm{X}$ is the candidate ``causers'' selected by pattern mining and $\mathrm{L}$ is the number of time lags. Thus the overall training time is $N_{iter}O(\mathrm{X}\mathrm{L} + (1+\mathrm{K})(\#))\mathrm{N}\mathrm{T}$. If the number of the graphical model (\#) is fixed, the computation time will approximately be at linear scalability with the sensor number $\mathrm{N}$ and timestamps number $\mathrm{T}$. We verified the linear scalability in Fig. \ref{fig:scalability}(b)(c). For the baseline methods,
$MCMC$ even cannot compute such large data sets. $CGBN$ and $K2+PS$ are unable to compute within 10 days and we leave their time cost as blank, as shown in \ref{fig:scalability}(c).
Meanwhile, the accuracy is guaranteed when extending city-level data to sensor-level data, as shown in Fig. \ref{fig:scalability}(a).
\begin{figure}[!htb]
\label{fig:scalability}

\centering
\includegraphics[width=3.3in]{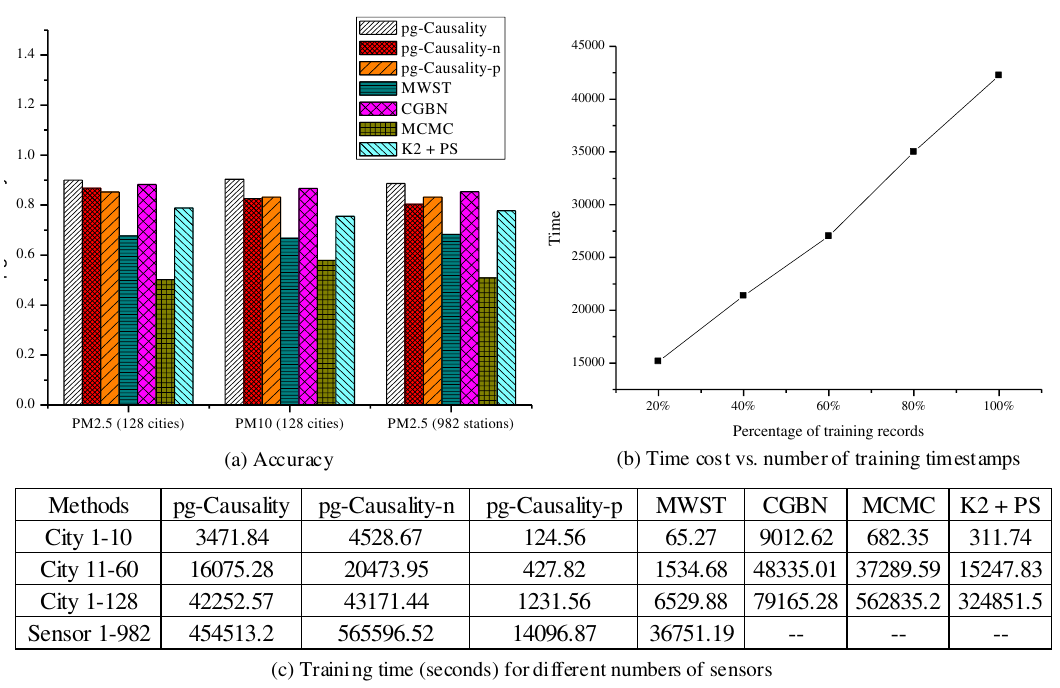}

\caption{\label{fig:scalability} Accuracy and time efficiency at city and station level.}

\end{figure}

\subsubsection{Verification with Synthetic Data}
Since the verification of causality via prediction task may not fully reflect the cause-and-effect relationships learned by the model, we further conduct experiments with synthetic data to judge whether the causality identification is correct or not.

As shown in Fig. \ref{fig:synthetic}, we generate N = 20 time series, with the pre-defined causal structure as in Fig. \ref{fig:synthetic}(a). This is done by randomly choosing the lag
k for any edge x $\rightarrow$ y in the feature causal graph \cite{arnold2007temporal}. To imitate the confounding effect, one time series is selected to influence all other time series. We reconstruct the causal structures through
Granger causality (as shown in Fig. \ref{fig:synthetic}(b)), lasso Granger causality (as shown in Fig. \ref{fig:synthetic}(c) \cite{arnold2007temporal}), and pg-Causality (as shown in Fig. \ref{fig:synthetic}(d)). To fit pg-Causality in this ``toy'' model, we simplified the model by randomly assigning locations to N time series. In the meanwhile, we set the distance constraint for selecting candidate ``causers'' to infinity, in order to consider every pair of causal relations between N time series. We mark the incorrect constructed edges in red. Result shows that pg-Causality generates the most likely structure compared with the baseline structure.

\begin{figure}[!htb]
\label{fig:synthetic}

\centering
\includegraphics[width=2.5in]{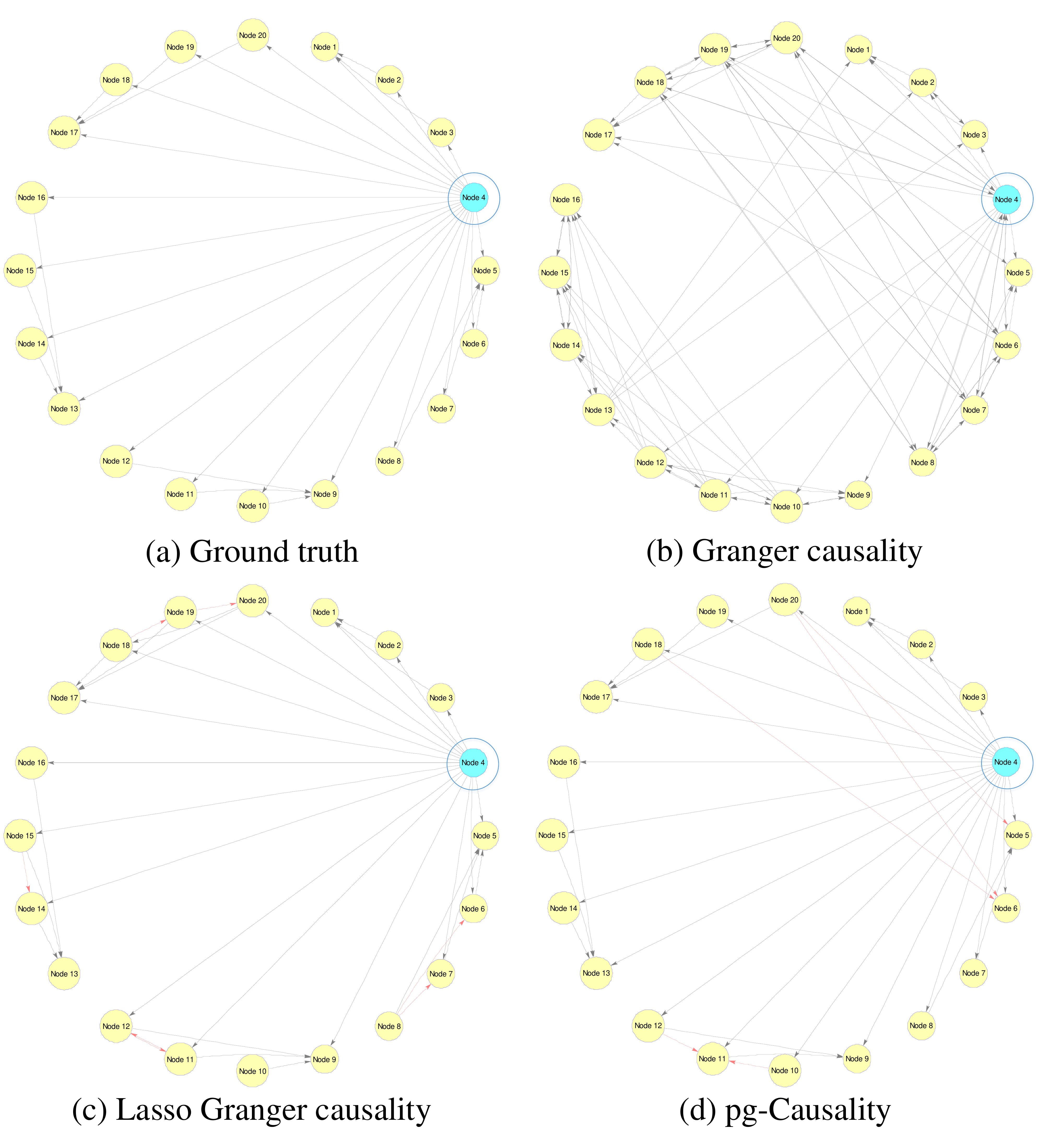}

\caption{\label{fig:synthetic} Causal structures generated by 20 synthetic time series. (a) Original structure with Node 4 (blue node, surrounded by a circle outside) as confounder, (b) Reconstructed by Granger causality, (c) Reconstructed by Lasso Granger causality, (d) Reconstructed by simplified pg-Causality. (Since the causa structure reconstructed by Granger causality in (b) significantly differs from the original one in (a), we only mark the incorrect connections for Lasso Granger causality and pg-Causality in red in (c) and (d).)}

\end{figure}

\subsubsection{Case Study}
\label{sect:casestudy}
To analyze the causal pathways for air pollutants, we study two cases corresponding to PM2.5 in specific cities. First we analyze the causal pathways for PM2.5 in the spring of Beijing and in the winter of Shanghai, the period of which are considered as the most heavily polluted season. Then we analyze Beijing PM2.5 before and during the APEC period (1st $-$ 14th, Nov, 2014) as a case study for human intervention in causal systems.

\pa{0ex}{1. Beijing and Shanghai.}
Fig. \ref{fig:training} is a real example for the causal pathways for Beijing PM2.5 during Jan$-$Mar. We provide the probability for each causal pathway for each cluster, defined as the proportion of labeled timestamps that belong to each cluster. As shown in Fig. \ref{fig:training}(a), Cluster 3 takes a relatively higher proportion (28.52\%) of time for Beijing PM2.5, indicating the causal pathway during Jan$-$Mar more probably come from southern sensors, i.e. Baoding and Cangzhou. Actions can be taken to control these pollutants in these cities. We then present the causal pathways for PM2.5 in Shanghai, during Oct$-$Dec, which statistically has the highest air pollution concentration. As shown in Fig. \ref{fig:visualization}, for PM2.5 in Shanghai, the $\mathrm{N}=3$ neighborhood cities generally come from the northwest and the southwest. Cluster 2 takes a relatively higher proportion (29.89\%) of time for Shanghai PM2.5, suggesting the pollutants may be dispersed from PM2.5 in Suzhou and Wuxi, and SO2 in Nantong.

\pa{0ex}{2. Beijing during APEC period.} 
Traditionally, causality is verified via interventions 
in a causal system. For example, we can verify the effect of a medicine by setting two groups of patients and only giving medicine to the treatment group. However, it is impossible to conduct intervention for air pollutants in the real environment. APEC period is a good opportunity to verify the causality, since the Chinese government shuttered factories in NC, and implemented traffic bans in and around Beijing \cite{huang2015apec}. Therefore, we compare the causal pathways for PM2.5 in Beijing before and during the APEC period. To illustrate the propagation of pollutants along the causal pathway, we connect the one-hop pathway to 3-hop as shown in Fig. \ref{fig:casestudy}(a)(b). The connection originates from the target pollutant, i.e., Beijing PM2.5, and connect its causal pollutants at neighbor cities. Then for each new connected pollutant, we repeat the same procedure for the next hop. The connection stops if in inference accuracy of one target pollutant based on its historical data is higher than based on the historical data of its ST ``causers'', indicating the pollutant is more likely to be generated locally. Fig. \ref{fig:casestudy}(a) shows Beijing's PM2.5 is likely to be caused by NO2 in Baoding (City 14), and PM10 in Cangzhou (City 18), during Jan $-$ Mar. Further, for example, Cangzhou's PM10 is mostly influenced by PM10 in Dingzhou (City 15) and Binzhou (City 71), as well as PM2.5 in Dezhou (City 64). We list the information of all 128 cities in Fig. \ref{fig:kn}, as well as their corresponding optimal K and N for pollutant PM2.5 in Spring. Note that the causal pathways forms ``circles'' in the southwestern cities to Beijing, which is identical to the locations of the major plants in NC shown in Fig. \ref{fig:casestudy}(c). However, we notice that the causal pathway cannot be connected into 3-hops during the APEC period, since each ``causer'' pollutant to Beijing PM2.5 (i.e. NO2 in Chengde and Zhangjiakou, and Tianjin) is more likely to be inferred by its own historical data over its ST ``causers'' in this period. This may suggest the PM2.5 in Beijing during the APEC period are mostly affected by pollutants locally and nearby. The 3-hop causal pathways learnt by three baselines are quite similar, thus we only present the result learned by CGBN, pg-Causality-p, pg-Causality-n, MWST, and MCMC in Fig. \ref{fig:casestudy}(d-h). Our approach has better interpretability. It is noted that without pattern mining module, the candidate ``causers'' for Beijing tend to be at irrelevant locations. While without integrating confounders, the causal pathways tend to have too many paths to be distinguished.
We summarize the discovery for
Beijing's air pollution as follows.

\begin{figure}[!htb]
\label{fig:visualization}

\centering
\includegraphics[width=3.4in]{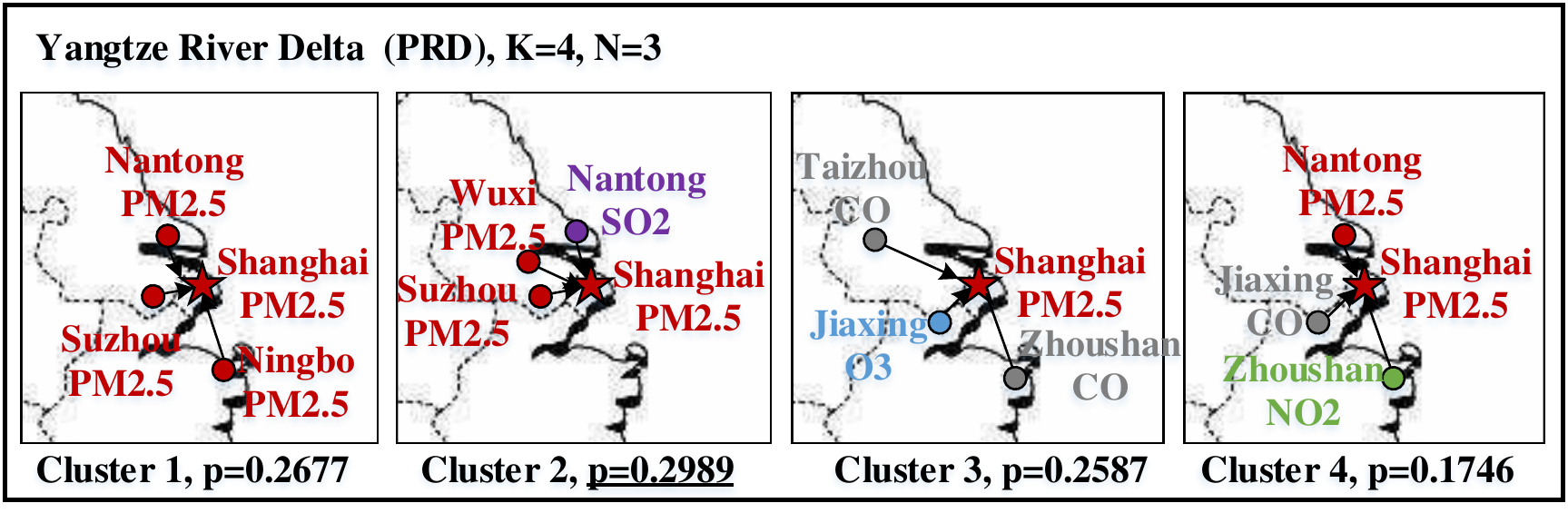}

\caption{\label{fig:visualization} Visualization of final causal pathways for PM2.5 in Shanghai.}

\end{figure}

\begin{figure*}[!htb]
\label{fig:casestudy}
\centering
\includegraphics[width=7in]{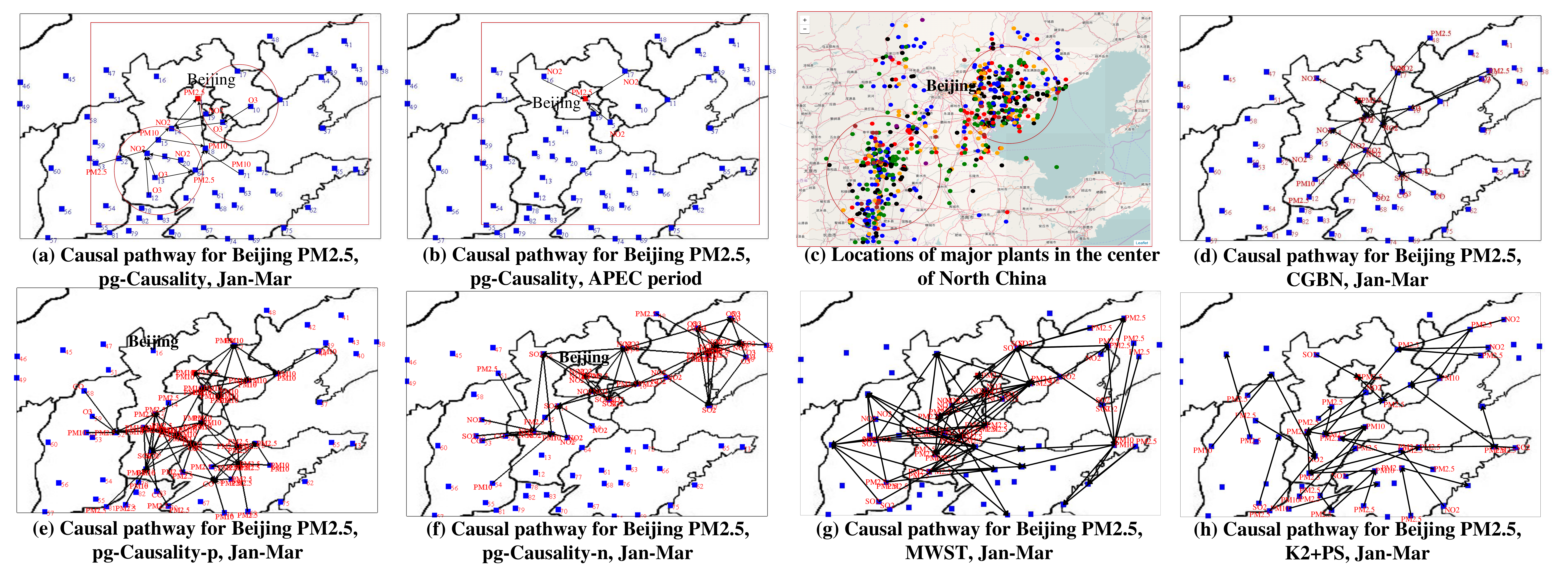}

\caption{\label{fig:casestudy} The causal pathways for Beijing PM2.5 before (a) and during APEC period (b), compared with the locations of major plants in Hebei Province, China (c), and the causal pathways learned by baseline method CGBN (d), pg-Causality-p (e), pg-Causality-n (f), MWST (g), MCMC (h).}

\end{figure*}

\begin{figure*}[!htb]
\label{fig:kn}
\centering
\includegraphics[width=7in]{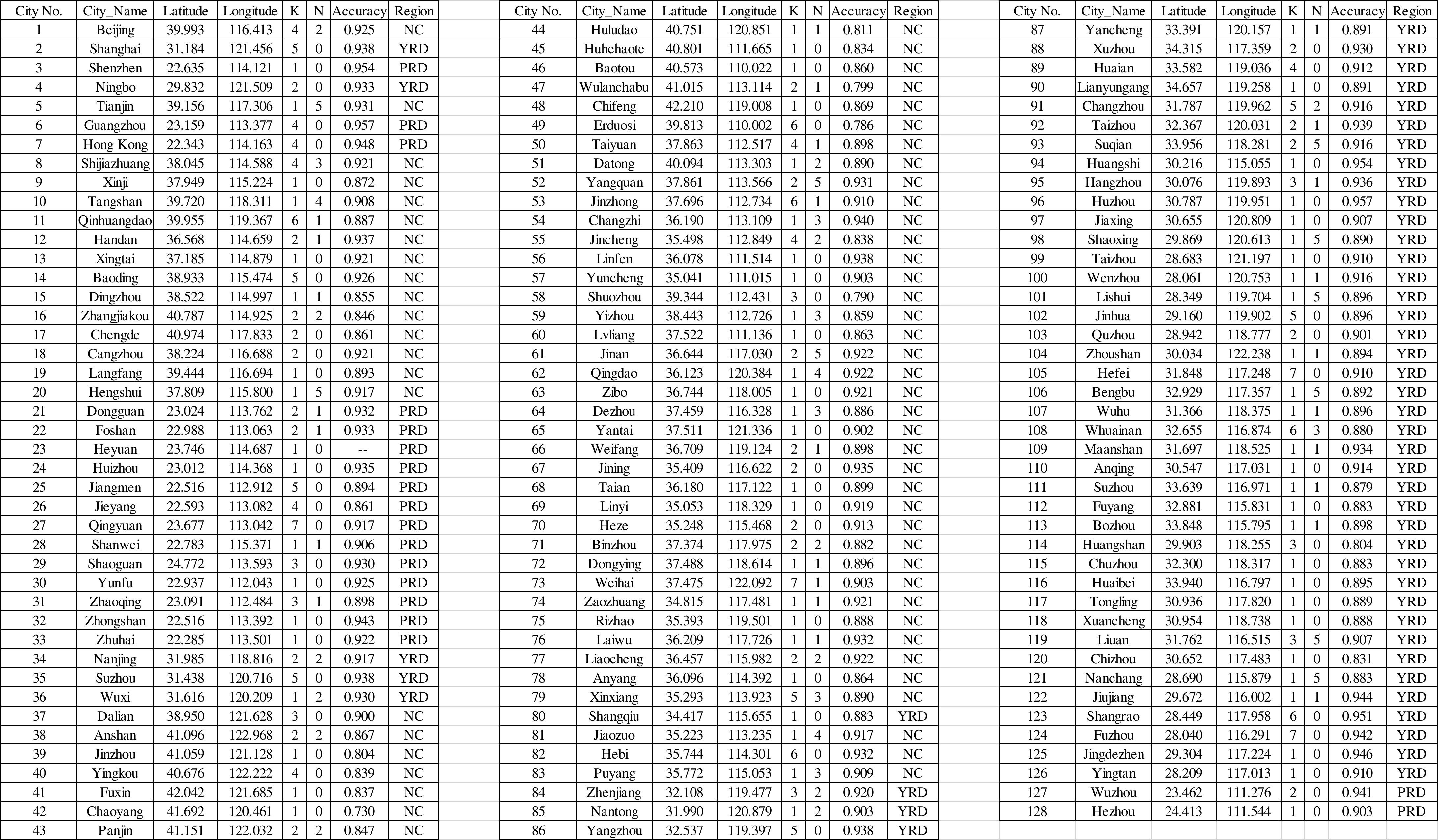}

\caption{\label{fig:kn} Optimal K and N for 128 cities, in Region NC, YRD and PRD, for PM2.5 during Jan $-$ Mar.}

\end{figure*}

$\bullet$ Among all the cities within a region, a target pollutant can be
 mainly affected by only several cities in the ST space. The
 locations of most influential cities to a target pollutant demonstrate
 seasonal similarities.

 $\bullet$ The causal pathways for PM2.5 in Beijing may come in multiple hops
 that form ``circle'' in the southwest of Beijing, suggesting superposition or reaction of air
 pollutants in the corresponding area. While during the APEC period with low pollution level, we did not see
 multi-hop causal pathways, suggesting the PM2.5 are more likely to be generated
 locally or nearby within this period.

\section{Related Work}
\label{sect:related}
\pa{0ex}{Data-driven Air Pollution Analysis:}
In recent years, air pollution analysis has drawn a lot of attention from the
data mining community ~\cite{zheng2014urban}\cite{zheng2015methodologies}. \cite{zheng2013u}\cite{zheng2015forecasting}\cite{hsieh2015inferring}
propose data-driven approaches to infer and forecast
fine-grained air quality using heterogeneous urban data.  \cite{shang2014inferring} estimates the gas consumption and pollutants emission of vehicles, based on the vehicles' GPS trajectories in the road network. Our paper differs
from these works in that, we target at understanding the underlying causal
pathways of air pollution. We identified the most likely ``causers'' in the geospace by learning the most likely graphical structures of an ST causality network, rather than predicting air quality or estimating
pollutant emission with a black-box neural network.

\pa{0ex}{Causality Modelling for Time Series:}
Causal modelling has been systematically studied for over half a century \cite{rubin1974estimating}\cite{granger1969investigating}, from the statistical and mathematical perspectives. 
For time
series data, existing works on modelling causality can be classified into three categories. The first category is based on Rubin's unit-level causality \cite{rubin1974estimating},
which is the statistical analysis on the potential outcome between two groups, given ``treatment'' and ``control'', respectively \cite{rosenbaum1983central}. 
With the increase of computation power, variations of unit-level causality were conducted, such as the cause-and-effect of advertising on behaviour change \cite{sun2015causal},
genes on phenotype \cite{wald2002homocysteine}, etc.
The second category considers
a pair of time series, and aims to quantify the strength of causal influence
from one time series to another.  Researchers have developed different measures
for this purpose, such as transfer entropy \cite{barnett2009granger}, and
Granger's causality~\cite{granger1969investigating}\cite{hlavavckova2007causality}.  The
third category aims to extract graphical causal relations from multiple time series.
\cite{arnold2007temporal} combines graphical
techniques with the classic Granger causality, and proposes a model to infer
causality strengths for a large number of time series variables. 
Pearl's causality model \cite{pearl2003causality} encodes the causal relationships in a directed acyclic graph (DAG)
\cite{heckerman1995learning} for probabilistic inference. The most well used graphical representation of DAG is Bayesian network (BN) \cite{heckerman1995learning}.
Temporal dependencies can be incorporated in the DAG by using Murphy's dynamic Bayesian network (DBN) \cite{murphy2002dynamic}.
There are also various extensions that incorporate
spatiotemporal dependencies in the domain of traffic \cite{liuwei2015traffic}, climate \cite{ebert2014causal}\cite{runge2012escaping}\cite{lozano2009spatial} and flood prediction \cite{jangyodsuk2015flood}. 

Our proposed approach \emph{pg-Causality} belongs to the third category, i.e., using
graphical model to detect causalities from multiple time series, where ``p'' refers to ``pattern-aided'' and ``g'' refers to graphical causality. The terms ``causality'' or ``causalities'' used later in this article are actually graphical causality.

The approach differs from the above works in three aspects: 
(1) As a data-driven causality learning method, we combine pattern mining and Bayesian learning to make the causality analysis more
efficient and robust to the noise present in the input data.
(2) Besides the multi-variate time series data, we also consider the impact of confounding given
different environmental factors for unbiased causality analysis. 
(3) Since we cannot conduct human intervention on air pollution at the nation-wide scale, this article identifies the causality from historical data. 
We proposed a Bayesian-based graphical causality model to capture the dependencies 
among different air pollution in the spatiotemporal (ST) space. Verification is based on the training accuracy, synthetic results, as well as observation.

\section{Conclusion}
\label{sect:conclusion}
In this paper, we identified the \emph{ST causal
pathways} for air pollutants using large-scale air quality data and meteorological
data. We have proposed a novel causal pathway learning approach named
\emph{pg-Causality} that tightly combines pattern mining and Bayesian
learning. Specifically, by extending existing sequential pattern mining
techniques,  \emph{pg-Causality} first extracts a set of FEPs for each sensor, which captures most regularities in the air polluting
process, largely suppresses data noise and reduces the complexity in the ST space. In the Bayesian learning module,
\emph{pg-Causality} leverages the pattern-matched data to train a graphical
structure, which carefully models multi-faceted causality and environmental
factors. We performed extensive experiments on three real-word data sets.
Experimental results demonstrate that the causal pathways detected by
\emph{pg-Causality} are highly interpretable and meaningful.  Moreover, it
outperforms baseline methods in both efficiency and inference
accuracy. For future work, we plan to apply this pattern-aided causality analysis framework for other tasks in the ST space, such as traffic
congestion analysis and human mobility modelling \cite{gmove}.






\bibliographystyle{abbrv}
\bibliography{cited}

\end{document}